\documentclass[10pt,journal,compsoc]{IEEEtran}
\ifCLASSOPTIONcompsoc
  \usepackage[nocompress]{cite}
\else
  \usepackage{cite}
\fi

\ifCLASSINFOpdf
\else
\fi

\usepackage{amsmath}
\usepackage{amssymb}

\usepackage{algorithmic}
\usepackage{algorithm}

\ifCLASSOPTIONcompsoc
  \usepackage[caption=false,font=footnotesize,labelfont=sf,textfont=sf]{subfig}
\else
  \usepackage[caption=false,font=footnotesize]{subfig}
\fi

\usepackage{amsfonts}
\usepackage{graphicx} 
\usepackage{epstopdf}
\usepackage{yhmath}
\usepackage{amscd}
\usepackage{latexsym}
\usepackage{setspace}
\usepackage{wrapfig} 
\usepackage{lipsum} 
\usepackage{mathrsfs}

\usepackage{color}
\usepackage{multirow}
\usepackage{tabu}
\usepackage{hhline}
\usepackage{arydshln}
\usepackage{booktabs}
\usepackage{tabularx}

\newcolumntype{C}{>{\centering\arraybackslash}X} 

\par
\begin{document}

%
\title{A Generic Multi-Projection-Center Model and Calibration Method for Light Field Cameras}
%
%
%
%


\author{Qi Zhang, Chunping Zhang, Jinbo Ling, Qing Wang,~\IEEEmembership{Member, IEEE} and Jingyi Yu 
\IEEEcompsocitemizethanks{\IEEEcompsocthanksitem Qi Zhang, Chunping Zhang, Jinbo Ling, and Qing Wang (corresponding author) are with the School of Computer Science, Northwestern Polytechnical University, Xi'an 710072, China (e-mail: qwang@nwpu.edu.cn). \IEEEcompsocthanksitem Jingyi Yu is with the ShanghaiTech University, Shanghai 200031, China, (e-mail: jingyi.udel@gmail.com)\protect\\
\IEEEcompsocthanksitem The work was supported by NSFC under Grant 61531014.}
\thanks{Manuscript received December 21, 2017; revised June 26, 2018.}} 

%
%

\markboth{IEEE TRANSACTIONS ON PATTERN ANALYSIS AND MACHINE INTELLIGENCE, VOL., No., }%
{ZHANG \MakeLowercase{\textit{\textit{ET AL.}}}: A GENERIC MULTI-PROJECTION-CENTER MODEL AND CALIBRATION METHOD FOR LIGHT FIELD CAMERAS}
%



\IEEEtitleabstractindextext{%
\begin{abstract}
Light field cameras can capture both spatial and angular information of light rays, enabling 3D reconstruction by a single exposure. The geometry of 3D reconstruction is affected by intrinsic parameters of a light field camera significantly. In the paper, we propose a multi-projection-center (MPC) model with 6 intrinsic parameters to characterize light field cameras based on traditional two-parallel-plane (TPP) representation. The MPC model can generally parameterize light field in different imaging formations, including conventional and focused light field cameras. By the constraints of 4D ray and 3D geometry, a 3D projective transformation is deduced to describe the relationship between geometric structure and the MPC coordinates. Based on the MPC model and projective transformation, we propose a calibration algorithm to verify our light field camera model. Our calibration method includes a close-form solution and a non-linear optimization by minimizing re-projection errors. Experimental results on both simulated and real scene data have verified the performance of our algorithm.
\end{abstract}

}

\maketitle

\IEEEdisplaynontitleabstractindextext

%
\IEEEpeerreviewmaketitle

\IEEEraisesectionheading{\section{Introduction}\label{sec:introduction}}

\IEEEPARstart{T}{he} micro-lens array (MLA) based light field cameras, including conventional light field camera \cite{ng2006digital} and focused light field camera \cite{lumsdaine2009focused}, can capture radiance information of light rays in both spatial and angular dimensions, \textit{i.e.}, 4D light field  \cite{levoy1996light,gortler1996lumigraph}. The data from light field camera is equivalent to narrow baseline images of traditional cameras with coplanar projection centers. The measurement of same point in multiple directions allows or strengthens the applications in computational photography and computer vision, such as digital refocusing \cite{ng2005fourier}, depth estimation \cite{Jeon2015Accurate}, segmentation \cite{Wanner2013Globally} and so on. Recent work also proposed the methods on light field registration \cite{johannsen2015linear} and stitching \cite{birklbauer2014panorama,guo2015enhancing} to expand the field of view (FOV). To support these applications, it is crucial to accurately calibrate light field cameras and establish exact relationship between the ray space and 3D scene.

It plays {an} important role to build a model for describing the ray sampling pattern of light field cameras. Previous approaches have dealt with imaging models on light field cameras in different optical designs \cite{dansereau2013decoding,johannsen2013calibration,bok2014geometric,hahne2014light,thomason2014calibration}. The common points are based on the fact that the micro-lens is regarded as a pinhole model and the main-lens is described as a thin-lens model. However, some of open issues still remain in the models and methods. Firstly, the proposed models focus on angular and spatial information of rays, but the relationship between light field and 3D scene geometry is not explored. Secondly, very little work has considered a generic model before to describe light field cameras with different image formations \cite{ng2006digital,lumsdaine2009focused}. Thirdly, existing intrinsic parameters of light field camera models are either redundant or incomplete such that corresponding solutions are neither effective nor efficient.

In the paper, we first propose a multi-projection-center (MPC) model based on two-parallel-plane (TPP) representation \cite{levoy1996light,gortler1996lumigraph}. Then we deduce the transformations between 3D scene geometry and 4D light rays. Based on geometry transformations in the MPC model, we characterize various light field cameras in a generic 6-intrinsic-parameter model and present an effective intrinsic parameter estimation algorithm. Experimental results on both virtual (simulated data) and physical (Lytro, Illum and a self-assembly focused) light field cameras have verified the effectiveness and efficiency of our model.

Our main contributions have three aspects, including

(1) We deduce the transformations to describe the relationship between light field and scene structure.

(2) We describe light field cameras with different image formations as a generic 6-parameter model without redundancy.

(3) We propose an effective intrinsic parameter estimation algorithm for light field cameras, including a closed-form linear solution and a nonlinear optimization.

The remainder of the paper is organized as follows. Section \ref{sec:related_work} summarizes related work on the models of light field cameras and calibration methods. Section \ref{sec:model} introduces our MPC model and the transformations between the 3D structure and 4D light field. Based on the theory of light field parameterization, a generic 6-intrinsic-parameter light field camera model is proposed. Section \ref{sec:verification} provides the details of our calibration method and analyzes computational complexity of the closed-form solution. In Section \ref{sec:exp}, we present extensive results on the simulated and real scene light fields, demonstrating more accurate intrinsic parameter estimation than previous work \cite{dansereau2013decoding,bok2014geometric}.

\section{Related Work}\label{sec:related_work}

To acquire 4D light field, there are various imaging systems developed from traditional camera. Wilburn \textit{et al.} \cite{wilburn2005high} present a camera array to obtain light field with high spatial and angular resolutions. Classic calibration approach is employed for the camera array \cite{vaish2004using}. More general, in traditional multi-view geometry framework, multiple cameras in different poses are defined as a set of unconstrained rays, which is known as Generalized Camera Model (GCM) \cite{pless2003using}. The ambiguity of the reconstructed scene is discussed in traditional topics \cite{hartley2003multiple}. However, such applications on the camera array are limited by its high cost and complex control. In contrast, the MLA enables a single camera to record 4D light field more conveniently and efficiently, though the baseline and spatial resolution are relatively smaller than camera array. Compared to the camera array, multiple projection centers of MLA-based light field camera are aligned on a plane strictly due to physical design. Recent work devotes to intrinsic parameter calibration of light field cameras in two designs \cite{ng2006digital,lumsdaine2009focused}, which are quite different according to the image pattern of micro-lenses.

The main difference of light field cameras is the relative position of main lens's imaging plane and the MLA plane \cite{Bishop2012The}. It determines rays' distribution from the same point, which affects the way to extract sub-apertures from raw image, \textit{i.e.}, the micro-lens images \cite{cho2013modeling,Wanner2011Generating}. However, the measurements of the same point in multiple directions are obtained in different types of light field cameras, equivalent to the data of GCM. Therefore, the light field camera model can use classic multi-view geometry theory for reference.

Recently, some state-of-the-art methods have proposed models on conventional light field camera, where multiple viewpoints or sub-apertures are convenient to be synthesized. Dansereau \textit{et al.} \cite{dansereau2013decoding} present a model to decode pixels into rays for a Lytro camera, where a 12-free-parameter transformation matrix is related to reference plane outside the camera (in nonlinear optimization, 10 intrinsic parameters and 5 distortion coefficients are finally estimated). However, the calibration method using traditional camera calibration algorithm is not effective, also there are redundant parameters in the decoding matrix. Bok \textit{et al.} \cite{bok2014geometric} formulate a geometric projection model consisting of a main lens and a MLA (their extended work has been published in IEEE TPAMI \cite{Bok2017PAMI}). Intrinsic parameters are estimated by conducting raw images directly and an analytical solution is deduced. Moreover, Thomason \textit{et al.} \cite{thomason2014calibration} try to deal with the misalignment of the MLA and estimated its position and orientation. 

Apart from this, other researchers have explored models on the focused light field camera, where multiple projections of the same point are convenient to be recognized. Johannsen \textit{et al.} \cite{johannsen2013calibration} propose to calibrate intrinsic parameters of the focused light field camera. By reconstructing 3D points from the parallax in adjacent micro-lens images, the parameters (including depth distortion) are estimated. However, the geometry center of micro image is on its micro-lens's optical axis in the camera model. This assumption causes inaccuracy on reconstructed points and estimated results are finally compensated by the coefficients of depth distortion. Hahne \textit{et al.} \cite{hahne2015refocusing} further discuss the influence of above-mentioned assumption, \textit{i.e.}, the deviation of micro-lens and its image. Heinze \textit{et al.} \cite{heinze2015automated} apply a similar model with Johannsen \textit{et al.} \cite{johannsen2013calibration} and deduce a linear initialization for intrinsic parameters.

In a word, previous light field camera models are either redundant or complex, which leads to a non-unique solution of intrinsic parameter estimation or inaccuracy of decoding light field. {An} unreliable camera model is also a bottleneck that might impede light field applications for computer vision and computational photography, especially on light field registration, stitching and enhancement. To support further applications,
a general light field camera model capable of representing rays and scene geometry more concisely is in urgent need.

\section{Multi-Projection-Center Model}\label{sec:model}
In this section, we first propose the MPC model based on the TPP representation of light field. Then we deduce the transformation matrix to relate 3D scene geometry and 4D rays. Finally, we utilize the MPC model to describe the image formation of light field cameras and define generic intrinsic parameters, including conventional and focused light field cameras. Table \ref{tab:term_notation} gives the notation of symbols used in the following sections.

\begin{table}[htbp]
\caption{Notation of symbols in the paper}
\centering
\footnotesize
\renewcommand\arraystretch{1}
\begin{tabular}{p{2.8cm}<{\centering}p{5.2cm}}
\toprule
Term & Definition\\
\midrule
$L(i,j,u,v)$ & Indexed pixel of raw image inside the camera\\
$L(I,J,U,V)$ & Virtual (conjugate) light field outside the camera\\
$L(s,t,x,y)$ & Decoded physical light field\\
$(k_i,k_j,k_u,k_v,u_0,v_0)$ & Intrinsic parameters\\
$\mathbf{X}_w$ & 3D point in the world coordinates \\
$\mathbf{X}_d$ & 3D point reconstructed by $L(i,j,u,v)$\\
$\mathbf{X}_c$ & 3D point reconstructed by $L(s,t,x,y)$\\
$\mathbf{R}_{3\!\times\!3}=[\mathbf{r}_1\,\,\mathbf{r}_2\,\,\mathbf{r}_3]$ & Rotation matrix of extrinsic parameter\\
$\mathbf{t}_{3\!\times\!1}=(t_x,t_y,t_z)^\top$ & Translation vector of extrinsic parameter\\
$\mathbf{M}_{2n\!\times\!4}$ & Measurement matrix of $n$ rays\\
$\mathbf{P}_{4\!\times\!4}$ & Homogenous projection matrix\\
$\mathbf{A}_{3\!\times\!3}$ & Non-homogenous projection matrix partitioned from $\mathbf{P}$\\
$\mathbf{H}_{4\!\times\!3}$ & Homography matrix decided by intrinsic and extrinsic parameters only\\
$\mathbf{d}\!=\!(k_1,k_2,k_3,k_4)^\top$ & Distortion vector\\
\bottomrule
\end{tabular}
\label{tab:term_notation}
\end{table}

\subsection{The Coordinates of MPC Model}\label{subsec:MPC}
\begin{figure}[tbp]
\centering
\includegraphics[width=2.8in]{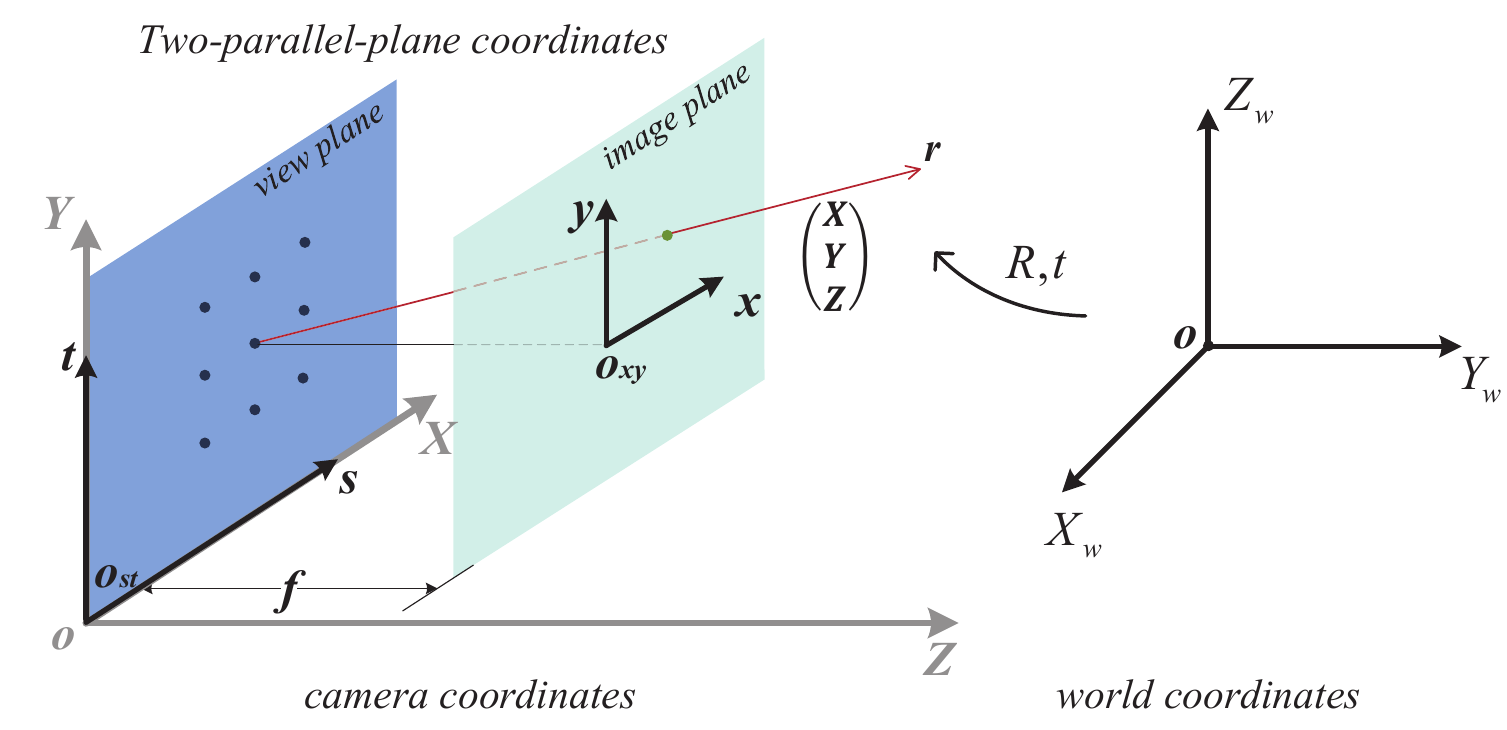}
\caption{An illustration of three coordinates in the MPC model.}
\label{fig:coordinate}
\end{figure}

As shown in Fig. \ref{fig:coordinate}, there are three coordinates in the MPC model, \textit{i.e.}, 3D world coordinates $O_wX_wY_wZ_w$, 3D camera coordinates $OXYZ$, 4D TPP coordinates $o_{st}st-o_{xy}xy$ ($o_{st}st$ for the view plane and $o_{xy}xy$ for the image plane). In general, the transformation between world and camera coordinates is related by extrinsic parameters $[\mathbf{R}|\mathbf{t}]$. The spacing between two parallel planes of traditional TPP representation is normalized as $1$ to describe a set of rays \cite{levoy1996light,gortler1996lumigraph}. Although it is complete and concise, to derive the transformation between 3D structure and 4D rays in light field cameras, we prefer a model consisting of two parallel planes with the spacing $f$. 

Let $L_f(s,t,x,y)$ denote light field in the MPC model with the spacing $f$. Then the ray is parameterized by two planes, {\it i.e.}, $Z\!=\!0$ and $Z\!=\!f$. Let $Z\!=\!0$ denote the view plane $o_{st}st$ and $Z\!=\!f$ denote the image plane $o_{xy}xy$. In the MPC model, $\mathbf{r}\!=\!(s,t,x,y)^\top$ defines a ray passing $(s,t,0)^\top$ and $(x,y, f)^\top$, where $(s,t,0)^\top$ is the projection center and $(x,y,f)^\top$ is the corresponding projection.

Given a projection center $(s,t,0)^\top$ (\textit{i.e.}, the $(s,t)$-$th$ view or sub-aperture) and the 3D point $\mathbf{X}\!=\!(X,Y, Z,1)^\top$, we can get the image projection $\mathbf{x}\!=\!(x,y,1)^\top$ in the local coordinate of the $(s,t)$-$th$ view,

\begin{equation}
 \lambda\left[ \begin{array}{c}
 x\\y\\1
 \end{array}   \right] = \left[   \begin{array}{cccc}
 f & 0 & 0 & -fs	\\
 0 & f & 0 & -ft	\\
 0 & 0 & 1 &  0
 \end{array} \right] \left[ \begin{array}{c}
 X \\ Y \\ Z \\ 1
 \end{array} \right].
  \label{eq:Ax=u}
\end{equation}

Since there are multiple projection centers $(s,t,0)^\top$, $s,t\!=\!1,2,\ldots,N$, the 3D point $\mathbf{X}$ can be observed for $N\!\times\!N$ times. Obviously, when the spacing $f$ changes to $1$ and there is only one projection center $(0,0,0)^\top$ on the view plane, the image formation degenerates into traditional central-projective camera model \cite{hartley2003multiple}.

\subsection{Transformation between Geometry and Rays}\label{subsec:coordinate}

It is known that different directional rays from one point enable 3D reconstruction. Let the ray $\mathbf{r}$ intersect at the point $\mathbf{X}$ in the 3D space, we can get the relationship between the ray and 3D point by the triangulation,

\begin{equation}
\underbrace{
  \left[  \begin{array}{cccc}
  f & 0 & -x & -fs \\
  0 & f & -y & -ft \end{array} \right] }_\mathbf{M}
  \left[ \begin{array} {c}
    X \\ Y \\ Z \\ 1 \end{array} \right] = \mathbf{0},
  \label{eq:MX=0_intersection}
\end{equation}

\noindent where $\mathbf{M}$ is a $2n\!\times\!4$ matrix consisting of $n$ rays and the MPC parameter $f$.

If two rays $\mathbf{r}_i\!=\!(s_i,t_i,x_i,y_i)^\top$ and $\mathbf{r}_j\!=\!(s_j,t_j,x_j,y_j)^\top$ are
from one 3D point $\mathbf{X}$, they can be represented by the following two equivalent forms,

\begin{equation}
\left[ \!\!\begin{array}{c}
X\\Y\\Z \end{array} \!\!\right]  \!= \frac{1}{x_i\!-\!x_j} \!\left[\!\! \begin{array}{c}
s_jx_i-s_ix_j\\ t_i(x_i\!-\!x_j)\!-\!y_i(s_i\!-\!s_j) \\ f(s_j\!-\!s_i) \end{array} \!\!\!\right]，
\label{eq:XYZ_form1}
\end{equation}

\noindent and

\begin{equation}
\left[ \!\!\begin{array}{c}
X\\Y\\Z \end{array} \!\!\right]  \!= \frac{1}{y_i\!-\!y_j}  \left[\!\! \begin{array}{c}
s_i(y_i\!-\!y_j)\!-\!x_i(t_i\!-\!t_j) \\ t_jy_i-t_iy_j \\ f(t_j\!-\!t_i) \end{array} \!\!\right].
\label{eq:XYZ_form2}
\end{equation}

\subsection{3D Projective Transformation}\label{subsec:3D_projective}

\begin{figure}[tbp]
\centering
\includegraphics[width=3.2in]{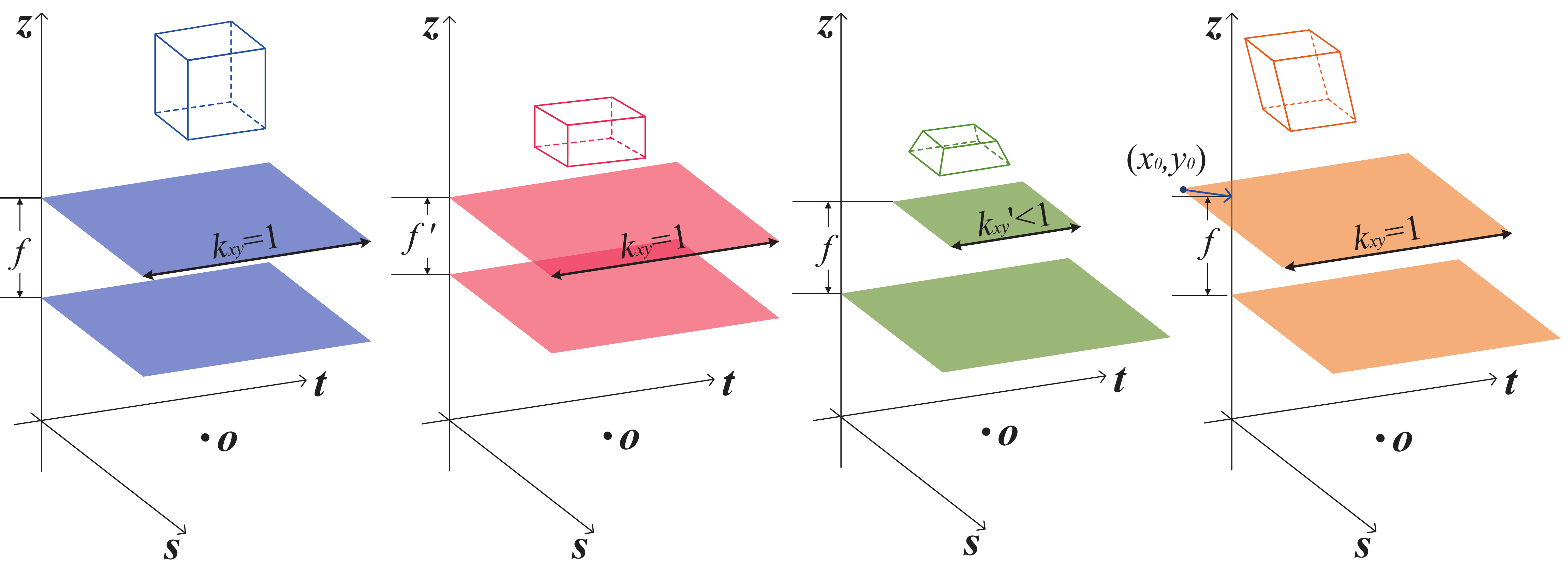}
\caption{Examples of transformations between 3D structure and 4D light field of a Lambertian cube. The leftmost is an original cube and others are the projected ones with the changing of parameter $f$, scaling in the image plane $k_{xy}$ ($=\!\!k_x\!\!=\!\!k_y$) and translation $(0,0,x_0,y_0)^\top$ respectively.}
\label{fig:3D_projective}
\end{figure}

In fact, a linear transformation on the coordinates of $\mathbf{r}$ causes 3D projective distortion on the reconstructed point $\mathbf{X}$ \cite{hartley2003multiple}, deduced from Eqs.(\ref{eq:XYZ_form1}) and (\ref{eq:XYZ_form2}). As shown in Fig. \ref{fig:3D_projective}, we show three examples of linear transformations, including the changing of $f$, scaling in the image plane $k_{xy}$ ($k_{xy}\!=\!k_x\!=\!k_y$) (in general there are 4 scaling factors $k_s,k_t,k_x,k_y$, two in the view plane and two in the image plane respectively), and translation in the image plane of specific view $(0,0,x_0,y_0)^\top$ (generally $(s_0,t_0,x_0,y_0)^\top$ in both planes).  The details are derived as follows.

(1) If we change $f$ into $f^\prime$, the imaging point $(x, y, f)^\top$ passed by $\mathbf{r}$ becomes $(x, y, f^\prime )^\top$ and the intersection of rays becomes $\mathbf{X}^\prime$. Substituting it into Eqs.(\ref{eq:XYZ_form1}) and (\ref{eq:XYZ_form2}), we have

\begin{equation}
\mathbf{X}^\prime = \mathbf{P}_1(f^\prime) \mathbf{X} = \left[  \begin{array}{cccc}
1 & 0 & 0       & 0 \\
0 & 1 & 0       & 0 \\
0 & 0 & f^\prime/f & 0 \\
0 & 0 & 0     & 1 \end{array}  \right] \left[  \begin{array}{c}
X \\ Y \\ Z \\ 1 \end{array}  \right],
\label{eq:P_f}
\end{equation}

\noindent where $\mathbf{X}$ and $\mathbf{X}^\prime$ are in the homogeneous coordinates.

(2) Let $\mathbf{r}$ become $\mathbf{r}^\prime\!=\!(s\!+\!s_0, t\!+\!t_0, x\!+\!x_0, y\!+\!y_0)^\top$, thus there is a transformation on the rays caused by the offset $\mathbf{m}=(s_0, t_0, x_0, y_0)^\top$. Substituting it into Eqs.(\ref{eq:XYZ_form1}) and (\ref{eq:XYZ_form2}), we can get the transformation matrix between $\mathbf{X}$ and $\mathbf{X}^\prime$,

\begin{equation}
\mathbf{X}^\prime = \mathbf{P}_2(\mathbf{m}) \mathbf{X} = \left[\!  \begin{array}{cccc}
1 & 0 & x_0/f & s_0 \\
0 & 1 & y_0/f & t_0 \\
0 & 0 & 1           & 0 \\
0 & 0 & 0           & 1 \end{array}  \!\right] \left[\!\!  \begin{array}{c}
X \\ Y \\ Z \\ 1 \end{array}  \!\!\right].
\label{eq:P_move}
\end{equation}

(3) Let $\mathbf{r}$ become $\mathbf{r}^\prime \!=\! (k_ss, k_tt, k_xx, k_yy)^\top$, thus there is a transformation caused by the scaling vector $\mathbf{k}\!=\!( k_s,k_t,k_x,k_y)^\top$. Then the transformation matrix between $\mathbf{X}$ and $\mathbf{X}^\prime$ is,

\begin{equation}
\mathbf{X}^\prime \!=\! \mathbf{P}_3(\mathbf{k})  \mathbf{X} \!=\!\! \left[\!\!\!  \begin{array}{cccc}
k_s & 0       & 0       & 0    \\
0       & k_t & 0       & 0    \\
0       & 0       & k_s/k_x    & 0    \\
0       & 0       & 0       & 1 \end{array}  \!\!\!\right]\!\!\!  \left[\!\!\!  \begin{array}{c}
X \\ Y \\ Z \\ 1 \end{array}  \!\!\!\right],
\label{eq:P_scale1}
\end{equation}

\noindent and

\begin{equation}
 \mathbf{X}^\prime \!=\! \mathbf{P}_4(\mathbf{k})  \mathbf{X} \!=\!\! \left[\!\!\!  \begin{array}{cccc}
k_s & 0       & 0       & 0    \\
0       & k_t & 0       & 0    \\
0       & 0       & k_t/k_y    & 0    \\
0       & 0       & 0       & 1 \end{array}  \!\!\!\right]\!\!\!  \left[\!\!\!  \begin{array}{c}
X \\ Y \\ Z \\ 1 \end{array}  \!\!\!\right].
\label{eq:P_scale2}
\end{equation}

\noindent In particular, Eqs.(\ref{eq:P_scale1}) and (\ref{eq:P_scale2}) hold when $k_s/k_t\!=\!k_x/k_y$.

As shown in the left-most of Fig. \ref{fig:3D_projective}, there is a scene with a Lambertian cube recorded by a MPC model. The observation of the cube in multiple directions is 4D light field. If the coordinates are linearly transformed and the light intensity keeps constant, the intersections of rays will be transformed by a 3D projection matrix. Therefore, the cube will be projected by transformation parameters (the right three of Fig. \ref{fig:3D_projective}).

\subsection{The MPC Model in Light Field Cameras}\label{subsec:MPC_camera}

Light field cameras are improved from traditional cameras. They record real world scene in different but similar ways. In traditional cameras, the central projection process of a 2D image is a dimension reduction of 3D space \cite{hartley2003multiple}. In light field camera, 3D structure projected by the main lens is arranged by the design of light path on the image sensor. The processes of multiple center projections are analyzed as follows.

\begin{figure}[bp]
\centering
\includegraphics[width=3.39in]{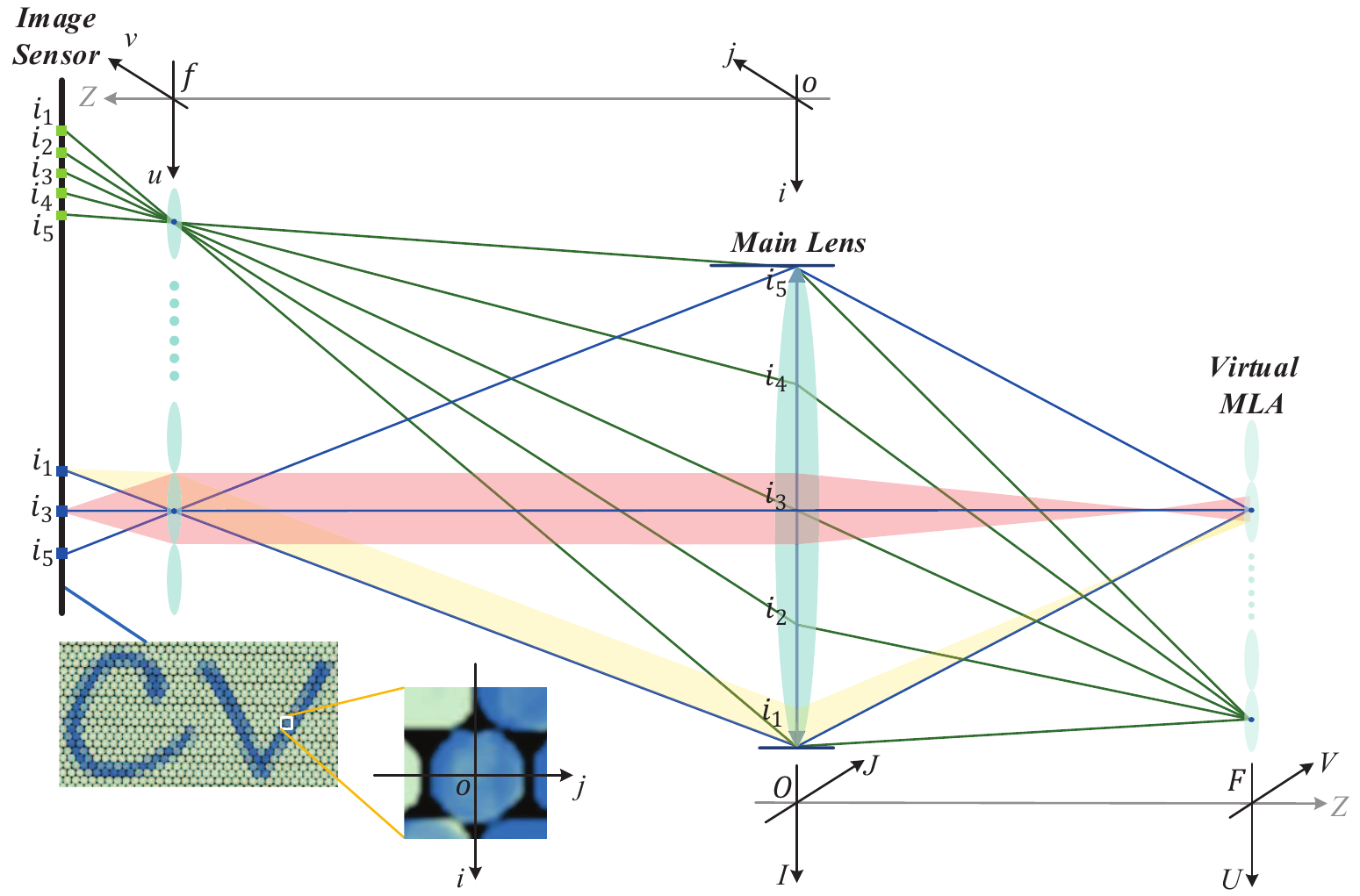}
\caption{Optical path of a conventional light field camera \cite{ng2006digital}. There are two MPC coordinates inside the camera and in the outer world with linear transformation, {\it i.e.}, $L_f(i,j,u,v)$ and $L_F(I,J,U,V)$ respectively.}
\label{fig:plenoptic1}
\end{figure}

On the one hand, as for a conventional light field camera, the sampling pattern of light field is shown in Fig. \ref{fig:plenoptic1}. The pixel $(u,v)^\top$ of $N\!\times\!N$ sub-aperture images is extracted from the micro-lens of $(u,v)^\top$. The sub-aperture image of the $(i,j)$ view is extracted from the pixels $(i,j)^\top$ in the local micro-lens image coordinates, as shown in Fig. \ref{fig:plenoptic1}. Obviously, there are two light fields, {\it i.e.}, $L_f(i,j,u,v)$ inside the camera and $L_F(I,J,U,V)$ in the outer world. Considering the projection of main lens, there is a 3D projective distortion between the 3D points reconstructed from $L_f$ and $L_F$.

\begin{figure}[tbp]
\includegraphics[width=3.3in]{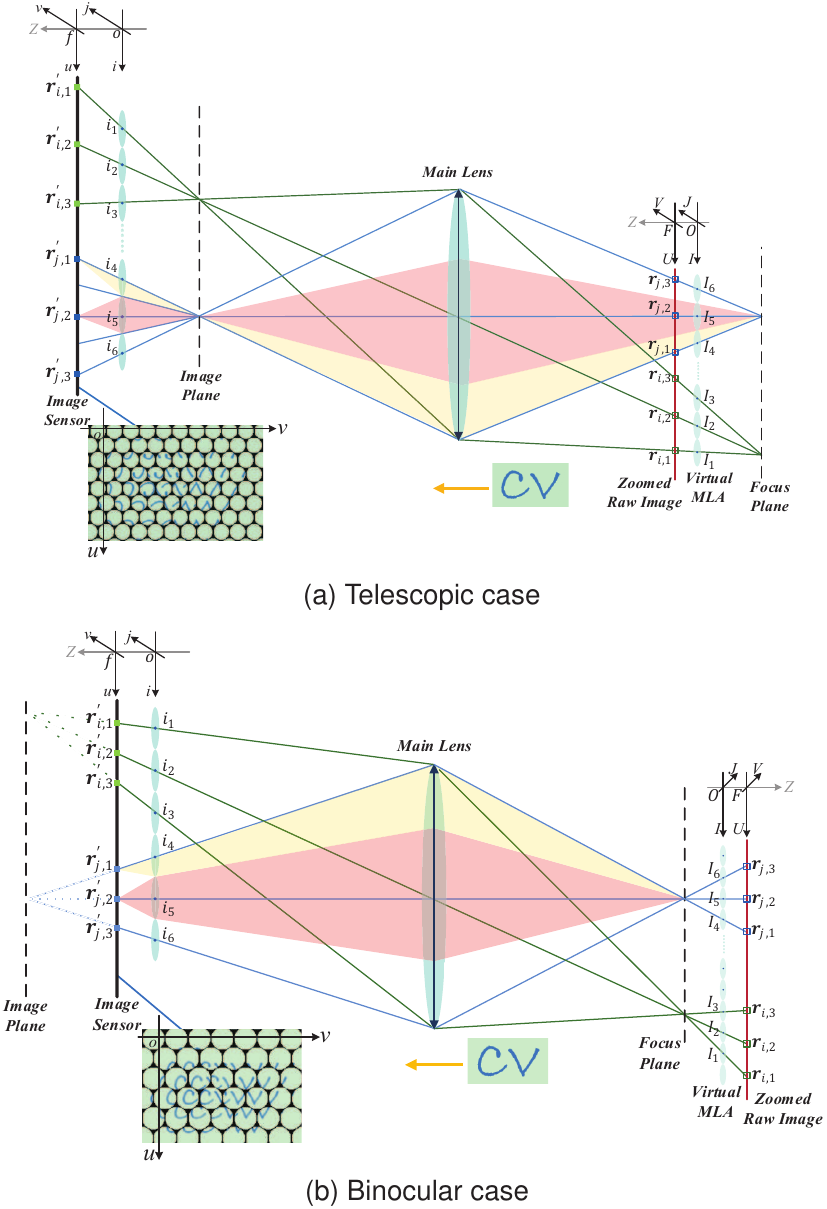}
\caption{Optical paths of focused light field cameras with different designs \cite{lumsdaine2009focused}. There is a conjugate MPC coordinate $L_F(I,J,U,V)$ in the outer world with the inner one $L_f(i,j,u,v)$.}
\label{fig:plenoptic2}
\end{figure}

On the other hand, as for the focused light field cameras, two sampling patterns of light field in two different optical paths are shown in Fig. \ref{fig:plenoptic2}. The micro-lenses project the distorted 3D scene inside the camera on the image sensor, where the image range is controlled by the aperture of main lens and the distance of components. The light field inside the camera can be decoded by the pixels of image sensor $(u,v,f)^\top$ and their corresponding optical centers of micro-lens $(i,j,0)^\top$, {\it i.e.}, $L_f(i,j,u,v)$. In addition, $(i,j)^\top$ is determined by the layout of MLA, as shown in Fig. 5b. By the transformation on the coordinate of $L_f(i,j,u,v)$ we have discussed in Sec \ref{subsec:3D_projective}, the outside light field $L_F(I,J,U,V)$ is obtained, which is the conjugate MPC coordinate outside the camera. The real world scene can be reconstructed by the light field $L_F(I,J,U,V)$ without projective distortion.

\begin{figure}[tbp]
\centering
\includegraphics[width=3.39in]{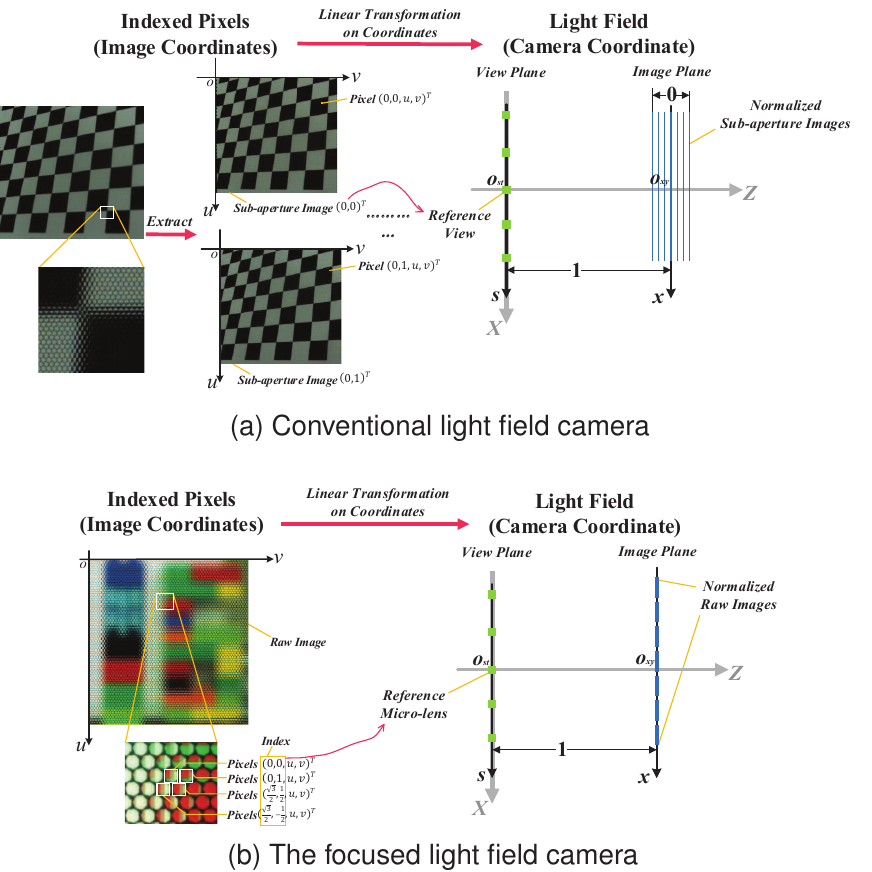}
\caption{The indexed pixels $L(i,j,u,v)$ and decoded physical light field $L(s,t,x,y)$ of light field cameras in two designs.}
\label{fig:cam_coor}
\end{figure}

Let $L(i,j,u,v)$ denote indexed pixels of light field cameras with $f\!=\!1$. Moreover, $L(i,j,u,v)$ is a set of indexed pixels and not a physical light field. In conventional light field camera, $L(i,j,u,v)$ are the sub-aperture images indexed by the $(i,j)$ view. In the focused light field cameras, $L(i,j,u,v)$ are micro-lens images indexed by their relative positions on the raw image. Obviously, by a linear transformation on the $L(i,j,u,v)$, we can conduct $L_F(I,J,U,V)$ and eliminate 3D projective distortion caused by the main lens. However, to parameterize 4D light field without redundancy, the spacing of two parallel planes should be 1. Let $L(s,t,x,y)$ denote the normalized light field. According to Eqs.(\ref{eq:P_f}) to (\ref{eq:P_scale2}), the normalization is a linear operation on the coordinates, and transformation matrices $\mathbf{P}_1$, $\mathbf{P}_2$ and $\mathbf{P}_3$ are all identity matrices. It means that indexed pixels $L(i,j,u,v)$ can be transformed to physical rays $L(s,t,x,y)$ in real world scene by linear transformations {as} we discussed before. The indexed pixels $L(i,j,u,v)$ and decoded physical light field $L(s,t,x,y)$ of light field cameras in two different designs are shown in Fig. \ref{fig:cam_coor}, where pixels and physical rays are related by intrinsic parameters.

In summary, we can transform an indexed pixel of raw image $L(i,j,u,v)$ into a normalized physical light field $L(s,t,x,y)$ by a decoding matrix $\mathbf{D}$ that is consisting of intrinsic parameters $(k_i,k_j,k_u,k_v,u_0,v_0)$.

\begin{equation}
\left[ \begin{array}{c}
s \\ t \\ x \\ y  \\ 1
\end{array}  \right] = \underbrace{
\left[  \begin{array}{ccccc}
k_i & 0   & 0   & 0   & 0   \\
0   & k_j & 0   & 0   & 0   \\
0   & 0   & k_u & 0   & u_0 \\
0   & 0   & 0   & k_v & v_0 \\
0   & 0   & 0   & 0   & 1   \end{array} \right]}_{=:\,\mathbf{D}}
\left[ \begin{array}{c}
i \\ j \\ u \\ v  \\ 1
\end{array} \right].
\label{eq:Decoding}
\end{equation}

Let $\mathbf{X}_d\!\!=\!\![X_d,Y_d,Z_d,1]^\top$ and $\mathbf{X}_c\!\!=\!\![X_c,Y_c,Z_c,1]^\top$ denote two 3D points reconstructed by $L(i,j,u,v)$ and $L(s,t,x,y)$ respectively. According to Eq.(\ref{eq:Decoding}), the relationship between $\mathbf{X}_d$ and $\mathbf{X}_c$ is

\begin{equation}
\underbrace{
\left[\!\!  \begin{array}{cccc}
1/k_i & 0     & -u_0/k_i & 0   \\
0     & 1/k_j & -v_0/k_j & 0   \\
0     & 0     & k_u/k_i  & 0   \\
0     & 0     & 0        & 1 \end{array} \!\!\!\right] }_{ =: \mathbf{P} }
\!\!\left[\!\!\! \begin{array}{c}
X_c \\ Y_c \\ Z_c \\ 1	
\end{array} \!\!\right] \!\!=\!\! \left[\!\!\! \begin{array}{c}
X_d \\ Y_d \\ Z_d \\ 1	
\end{array} \!\!\right],
\label{eq:PXc=Xd}
\end{equation}	

\noindent where $\mathbf{P}\!\!=\!\!\mathbf{P}_{1}(1)\mathbf{P}_{3}(\mathbf{k})\mathbf{P}_{2}(\mathbf{m})$ is determined by intrinsic parameters in the decoding matrix $\mathbf{D}$. Here, $\mathbf{m}=(0,0,-u_0,-v_0)^\top$ and $\mathbf{k}=(1/k_i,1/k_j,1/k_u,1/k_v)^\top$, which are totally decided by the mapping from indexed pixels to real world light rays.

In addition, the light field inside a conventional light field camera (in Fig. \ref{fig:plenoptic1}) can also be parameterized by the MPC model that is consisting of image sensor and the MLA. However, considering the convenience of extracting sub-aperture images and the difficulty on detecting points on raw image in a conventional light field camera, we prefer to discuss the data as a set of sub-aperture images. Conversely, for the focused one, we model the parameterization plane by the raw image plane and discuss the raw image directly.


\section{Light Field Camera Calibration}\label{sec:verification}

We verify our light field camera model by intrinsic parameter calibration. We will provide the details of how to solve intrinsic parameters, including a linear closed-form solution and a nonlinear optimization to minimize the re-projection error. In our method, the prior scene points are supported by a planar calibration board in different poses.

\subsection{Linear Initialization}\label{subsec:linear}

After necessary preprocessing, the micro-lens images are recognized \cite{dansereau2013decoding,cho2013modeling,Chunping2016Decoding}, {\it i.e.}, $L(i, j, u, v)$. We assume that the prior 3D point $\mathbf{X}_w$ in the world coordinates is related to the 3D point $\mathbf{X}_c$ in the MPC coordinates by a rigid motion, $\mathbf{X}_c\!=\!\mathbf{R}\mathbf{X}_w + \mathbf{t}$, with the rotation $\mathbf{R} \in SO(3)$ and translation $\mathbf{t}\!=\!(t_x,t_y,t_z)^\top \in \mathbb{R}^3 $. Let $\mathbf{r}_i$ denote $i$-$th$ column vector of $\mathbf{R}$. The relationship among $\mathbf{R}$, $\mathbf{t}$, $\mathbf{X}_w$ and intrinsic parameters $\mathcal{P}\!=\!(k_i,k_j,k_u,k_v,u_0,v_0)$ is obtained by Eqs.(\ref{eq:MX=0_intersection}) and (\ref{eq:PXc=Xd}).

\begin{equation}
\mathbf{M} \mathbf{P} \left[ \begin{array} {cccc}
\mathbf{r}_1 & \mathbf{r}_2 & \mathbf{r}_3 & \mathbf{t} \\
0        & 0        & 0        & 1
\end{array} \right] \mathbf{X}_w = \mathbf{0}.
\label{eq:MP[Rt]Xw=0}
\end{equation}
\noindent where $\mathbf{M}$ is a $2n\!\times\!4$ measurement matrix of $n$ rays and $n\!\ge\!2$. These rays are derived from the indexed pixels $L(i,j,u,v)$ as mentioned in Eq.(\ref{eq:Ax=u}).

Suppose that the calibration board is on the plane of $Z\!=\!0$ in the world coordinates, thus $Z_w=0$. To solve the unknown parameters, we simplify Eq.(\ref{eq:MP[Rt]Xw=0}) as,

\begin{equation}
\mathbf{M} \otimes \left[ X_w \,\,\, Y_w \,\,\, 1 \right] \vec{\mathbf{H}}  = \mathbf{0},
\label{eq:MXwH=0}
\end{equation}

\noindent where $\vec{\mathbf{H}}$ is a $12\!\times\!1$ matrix stretched on row from $\mathbf{H}$. $\otimes$ is a direct product operator. $\mathbf{H}$ is a $4\!\times\!3$ matrix only consisting of intrinsic and extrinsic parameters, defined as

\begin{equation}
\mathbf{H}=\mathbf{P}\left[ \begin{array}{ccc}
   \mathbf{r}_1 & \mathbf{r}_2 & \mathbf{t}  \\
   0 & 0 & 1
\end{array} \right].
\label{eq:H=Prt}	
\end{equation}

In addition, $\mathbf{M}$ is a matrix containing at least 2 rays from light field $L(i,j,u,v)$, according to Eq.(\ref{eq:MX=0_intersection}). By stacking measurements from at least 3 non-collinear points $\mathbf{X}_w$, the homography $\mathbf{H}$ can be estimated by Eq.(\ref{eq:MXwH=0}). 

In order to derive intrinsic parameters from $\mathbf{H}$, we can partition $\mathbf{P}$ to extract a $3\!\times\!3$ upper triangle matrix $\mathbf{A}$. Let $h_{ij}$ denote the element on the $i$-$th$ row and $j$-$th$ column of $\mathbf{H}$, we rewrite Eq.(\ref{eq:H=Prt}) as follows,

\begin{equation}
\left[ \begin{array}{c;{2pt/2pt}c}
             & h_{13} \\	
\mathbf{G}   & h_{23} \\	
             & h_{33} \\\hdashline
\mathbf{0}_{1\times2}  & 1
\end{array} \right]\!\! =\!\! \left[ \begin{array}{c;{2pt/2pt}c}
             & 0 \\	
\mathbf{A}   & 0 \\	
             & 0 \\ \hdashline
\mathbf{0}_{1\times3} & 1
\end{array} \right] \!\! \left[ \begin{array} {ccc}
\mathbf{r}_1 & \mathbf{r}_2 & \mathbf{t} \\
0        & 0        & 1
\end{array} \right],
\label{eq:H3x3=A[r1r2]}	
\end{equation}
\noindent where $\mathbf{G}$ is a $3\!\times\!2$ matrix, \textit{i.e.}, top-left $3\!\times\!2$ of $\mathbf{H}$.

Let $\mathbf{g}_j\!=\!(g_{1j},g_{2j},g_{3j})^\top, j\!=\!1,2$ denote the $j$-$th$ column vector of $\mathbf{G}$. Utilizing the orthogonality and identity of $\mathbf{R}$, we have

\begin{equation}
\begin{aligned}
&\mathbf{g}_1^{\top} \mathbf{A}^{-\top} \mathbf{A}^{-1} \mathbf{g}_2 = 0, \\
&\mathbf{g}_1^{\top} \mathbf{A}^{-\top} \mathbf{A}^{-1} \mathbf{g}_1 = \mathbf{g}_2^{\top} \mathbf{A}^{-\top} \mathbf{A}^{-1} \mathbf{g}_2.
\end{aligned}
\label{eq:giAAgj}
\end{equation}
\noindent where
$
\mathbf{A}^{-1} \!\!=\!\! \left[\!\!\! \begin{array} {ccc}
k_i    & 0    & u_0k_i/k_u  \\
0      & k_j  & v_0k_j/k_v  \\
0      & 0    & k_i/k_u
\end{array}	\!\!\!\right]
$.

Let a symmetric matrix $\mathbf{B}$ denote $\mathbf{A}^{-\top} \mathbf{A}^{-1}$. The analytical form of $\mathbf{B}$ is

\begin{equation}
\mathbf{B} \!=\! \left[\!\! \begin{array} {ccc}   
k_i^2         & 0             & k_i^2u_0/k_u                     \\
0             & k_j^2         & k_j^2v_0/k_v                     \\
k_i^2u_0/k_u  & k_j^2v_0/k_v  & k_i^2/k_u^2\left(1\!+\!u_0^2\!+\!v_0^2\right)
\end{array}	\!\!\right].
\label{eq:B=AA}
\end{equation}

Note that there are only 5 distinct non-zero elements in $\mathbf{B}$, denoted by $\mathbf{b}\!:=\!(b_{11}, b_{13}, b_{22}, b_{23}, b_{33})^\top$. To solve $\mathbf{B}$, we rewrite Eq.(\ref{eq:giAAgj}) as follows,

\begin{equation}
\left[\!\! \begin{array}{cc}
g_{11}g_{12}              &  g_{11}^2-g_{12}^2\\
g_{11}g_{32}\!+\!g_{12}g_{31} &  2( g_{11}g_{31}\!-\!g_{12}g_{32})\\
g_{21}g_{22}              &  g_{21}^2-g_{22}^2\\
g_{21}g_{32}\!+\!g_{31}g_{22} &  2( g_{21}g_{31}\!-\!g_{22}g_{32})\\
g_{31}g_{32}              &  g_{31}^2-g_{32}^2
\end{array} \!\!\right]^{\!\!\top}
\mathbf{b}
\!= \mathbf{0}.
\label{eq:gb=0}
\end{equation}

By stacking at least two such equations (from two poses) as Eq.(\ref{eq:gb=0}), we can obtain a unique general non-zeros solution for $\mathbf{b}$, which is defined up to an unknown scale factor.

Once $\mathbf{B}=\mathbf{A}^{-\top} \mathbf{A}^{-1}$ is determined, it is an easy matter to solve $\mathbf{A}^{-1}$ using Cholesky factorization \cite{Hartley1997Self}. Let $\mathbf{\hat{A}}$ denote the estimation of $\mathbf{A}$, \textit{i.e.}, $\lambda \mathbf{\hat{A}}=\mathbf{A}$. Let $\hat{a}_{ij}$ denote the element on the $i$-$th$ row and $j$-$th$ column of $\mathbf{\hat{A}}$, intrinsic parameters except $k_i$ and $k_j$ are estimated by the ratio of elements

\begin{equation}
\begin{aligned}
& k_u = \hat{a}_{11}/\hat{a}_{33},
& k_v = \hat{a}_{22}/\hat{a}_{33}, \\
& u_0 = \hat{a}_{13}/\hat{a}_{33},
& v_0 = \hat{a}_{23}/\hat{a}_{33}.
\end{aligned}	
\label{eq:intrinsics}
\end{equation}

Apart from intrinsic parameters, extrinsic parameters in different poses can be extracted as follows,

\begin{equation}
\begin{aligned}
& \lambda \;= \alpha\left(\left\| \mathbf{\hat{A}}^{-1} \mathbf{g}_1 \right\| + \left\| \mathbf{\hat{A}}^{-1} \mathbf{g}_2 \right\| \right) / 2, \\
& \mathbf{r}_1 = \mathbf{\hat{A}}^{-1} \mathbf{g}_1 / \lambda,\\
& \mathbf{r}_2 = \mathbf{\hat{A}}^{-1} \mathbf{g}_2 / \lambda,\\
& \mathbf{t}\;\; = \mathbf{\hat{A}}^{-1} \left[ h_{13} \ h_{23} \  h_{33} \right]^\top / \lambda,\\
& \mathbf{r}_3 = \mathbf{r}_1 \times \mathbf{r}_2,\\
\end{aligned}
\label{eq:extrinsics}
\end{equation}

\noindent where $\|\!\cdot\!\|$ denotes $L_2$ norm. $\alpha$ values 1 or -1 and it is decided by image formation. In conventional light field camera and the focused one with shorter light path (as shown in Fig. \ref{fig:plenoptic1} and 4b), $\alpha$ makes $t_z > 0$. Otherwise, in the focused light field camera with longer light path (see Fig. 4a), $\alpha$ makes $t_z < 0$.

To obtain other two intrinsic parameters $k_i$ and $k_j$, we substitute the results in Eq.(\ref{eq:extrinsics}) for Eq.(\ref{eq:MP[Rt]Xw=0}) and obtain $\mathbf{X}_c\!=\!\mathbf{R}\mathbf{X}_w + \mathbf{t}$ using the estimated extrinsic parameters. Then, Eq.(\ref{eq:MX=0_intersection}) is rewritten as,

\begin{equation}
\left[ \!\begin{array} {cc}
i & 0 \\
0 & j
\end{array} \!\right]	\left[ \!\begin{array} {c}
k_i \\ k_j	
\end{array}\! \right] = \left[\! \begin{array} {c}
X_c\!-\!xZ_c \\ Y_c\!-\!yZ_c	
\end{array} \!\right].
\label{eq:solve_kikj}
\end{equation}

Stacking the measurements in different poses, we can obtain a unique non-zeros solution for $k_i$ and $k_j$.

\subsection{Nonlinear optimization}\label{subsec:non_linear}

The most common distortion of traditional camera is radial distortion. The optical property of main lens and physical machining error of the MLA might lead to the distortion of rays in light field camera. Theoretically, due to two level imaging design with main lens and micro-lens array, there should exist radial distortion on the image plane $(x,y)^\top$ and sampling distortion on the view plane $(s,t)^\top$ simultaneously. In the paper, we only consider the distortion on the image plane and omit sampling distortion on the view plane (\textit{i.e.}, angular sampling grid is ideal without distortion). 

\begin{equation}
\left\{
\begin{aligned}
&x^u = (1 + k_1r_{xy}^2 + k_2r_{xy}^4 )x + k_3s & \\	
&y^u = (1 + k_1r_{xy}^2 + k_2r_{xy}^4 )y + k_4t & \\	
\end{aligned}
\right.,
\label{eq:distortion}
\end{equation}

\noindent where $r_{xy}^2\!=\!{x^2+y^2}$ and $\mathbf{r}\!=\!(s,t,x,y)^\top$ is the ray transformed from the measurement $L(i,j,u,v)$ by intrinsic parameter $\mathcal{P}$ according to Eq.(\ref{eq:Decoding}). $\mathbf{d}\!=\!(k_1, k_2, k_3, k_4)^\top$ denotes distortion vector and $\mathbf{x}^u\!=\!(x^u, y^u)^\top$ is undistorted projection from the distorted one $\mathbf{x}\!=\!(x, y)^\top$ in the local image coordinates under the $(s,t)^\top$ view. In the distortion vector $\mathbf{d}$, $k_1$ and $k_2$ regulate radial distortion on the image plane. $k_3$ and $k_4$ represent the distortion of image plane affected by the sampling view $(s,t)^\top$, which is caused by non-paraxial rays of the main lens.

We minimize the following cost function with the initialization solved in Section \ref{subsec:linear} to refine the parameters, including intrinsic parameter $\mathcal{P}$, distortion vector $\mathbf{d}$, and extrinsic parameters $\mathbf{R}_p$ and $\mathbf{t}_p$, $p=1,\ldots,P$, $P$ is the number of poses.

\begin{equation}
\sum_{p=1}^{\#\!pose}\sum_{n=1}^{\#\!point}\sum_{i=1}^{\#\!view} { \left\| \mathbf{x}_i^u(\mathcal{P}, \mathbf{d}) - \hat{\mathbf{x}}_i (\mathbf{R}_p, \mathbf{t}_p, \mathbf{X}_{w,n} ) \right\| },
\label{eq:cost_function}
\end{equation}

\noindent where $\mathbf{x}^u$ is the image point from $L(i,j,u,v)$ according to Eq.(\ref{eq:Decoding}) and followed by distortion rectification according to Eq.(\ref{eq:distortion}). $\hat{\mathbf{x}}$ is the projection of 3D point $\mathbf{X}_{w,n}$ in the world coordinates according to Eq.(\ref{eq:Ax=u}).

In Eq.(\ref{eq:cost_function}), $\mathbf{R}$ is parameterized by Rodrigues formula \cite{faugeras1993three}. In addition, the Jacobian matrix of cost function is simple and sparse. This nonlinear minimization problem can be solved with the Levenberg-Marquardt algorithm based on trust region method \cite{madsen2004methods}. We adopt MATLAB's ${\mathsf{lsqnonlin}}$ function to complete the optimization. 

\begin{algorithm}[htb]
\small
  \begin{algorithmic}[1]
    \REQUIRE
     3D prior points $\mathbf{X}_w$ and corresponding rays $L(i,j,u,v)$.\\
    \ENSURE
    Intrinsic parameters $\mathcal{P}=(k_i,k_j,k_u,k_v,u_0,v_0)$;\\
    Extrinsic parameters $\mathbf{R}_p$, $\mathbf{t}_p(1\!\le\!p\!\le\!P)$;\\
    Distortion vector $\mathbf{d}=(k_1,k_2,k_3,k_4)^\top$.\\
    \FOR {$p=1$ to $P$}
    	\FOR{each 3D point $\mathbf{X}_w$}
    		\STATE{Generate the measurement matrix $\mathbf{M}$ from indexed pixel $L(i,j,u,v)$ \hfill $\triangleright$ Eq.(\ref{eq:MX=0_intersection})}
    	\ENDFOR
    	\STATE{Calculate the homography matrix $\mathbf{H}_p$ according to $\mathbf{X}_w$ and $\mathbf{M}$  \hfill $\triangleright$ Eq.(\ref{eq:MXwH=0})}
    \ENDFOR
    	\STATE{Calculate the matrix $\hat{\mathbf{B}}$ \hfill $\triangleright$ Eq.(\ref{eq:gb=0})}
    \STATE{Calculate projection matrix $\hat{\mathbf{A}}$ from $\hat{\mathbf{B}}$ using Cholesky factorization}
    \STATE{Obtain four intrinsic parameters $(k_u,k_v,u_0,v_0)$ \hfill $\triangleright$ Eq.(\ref{eq:intrinsics})}
    \FOR {$p=1$ to $P$}
        \STATE{Get extrinsic parameters $\mathbf{R}_p$ and $\mathbf{t}_p$ \hfill $\triangleright$ Eq.(\ref{eq:extrinsics})}
   	\ENDFOR
   	\STATE {Obtain other two intrinsic parameters $(k_i,k_j)$ \hfill $\triangleright$ Eq.(\ref{eq:solve_kikj})}
   	\STATE {Initialize distortion coefficient $\mathbf{d}=(0,0,0,0)^\top$}
    \STATE {Create the cost function according to intrinsic parameters, extrinsic parameters and distortion coefficient \hfill $\triangleright$ Eq.(\ref{eq:cost_function})}
    \STATE {Obtain optimized results using nonlinear LM algorithm}
  \end{algorithmic}
  \caption{Light Field Camera Calibration Algorithm.}
\label{alg:calibration_alg}
\end{algorithm}

\subsection{Computational Complexity}\label{subsec:complexity}
The calibration algorithm of light field camera is summarized in Alg.~\ref{alg:calibration_alg}. Let $(S,T)$ denote sampling number on the view plane, $(M,N)$ be the number of prior points on the calibration board, and $P$ be the number of poses, respectively. For the measurement of each pose, there are $(2\!\times\!S\times\!T)\times\!12$ linear equations to solve $\mathbf{H}$. Then $(2\!\times\!P)\times\!5$ linear equations and $(2\!\times\!P\times\!M\times\!N)\times\!2$ equations are solved to obtain intrinsic parameters. The main complexity is spent on the solution of $\mathbf{H}$ from different poses, \textit{i.e.}, $O(P)$.

By contrast, the algorithm in Dansereau \textit{et al.} \cite{dansereau2013decoding} calculates a homography for every sub-aperture {image} in different view, \textit{i.e.}, $O(P\!\times\!S\!\times\!T)$. It suffers from a higher complexity and a lower accuracy on parameter initialization. The algorithm in Bok \textit{et al.} \cite{bok2014geometric} solves a linear equation on every pose and its computational complexity is $O(P)$. However, there are intrinsic and extrinsic parameters in the equations, which causes inaccuracy on the solution. 

\section{Experimental Results}\label{sec:exp}
In this section, we verify our light field camera model by the calibration of intrinsic parameters. We present various experimental results both on simulated and real datasets. The performance is analyzed by comparing with the ground truth or baseline algorithms \cite{dansereau2013decoding} and \cite{bok2014geometric}.

\subsection{Simulated data}\label{subsec:simu}

In this subsection we verify our calibration method on simulated data. The simulated light field camera has the following property referred to Eq.(\ref{eq:Decoding}), as shown in Table \ref{tab:configuration}. These parameters are close to the setting of Lytro camera so that we obtain plausible input close to real-world scenarios. The checkerboard is a pattern with $12\!\times\!12$ points with $3.51mm\!\times\!3.51mm$ cells.

\subsubsection{Performance w.r.t the number of poses and views}
Firstly, we test the performance with respect to the number of poses and the number of views. We vary the number of poses from 2 to 8 and the number of views from $2\!\times\!2$ to $7\!\times\!7$. For each combination of pose and view, 200 trails of independent calibration board poses are generated. The rotation angles are randomly generated from $-30^\circ$ to $30^\circ$, and the measurements are all added with Gaussian noise with zero mean and standard deviation 0.5 pixels.

\begin{figure}[!tbp]
\centering
\includegraphics[width=3.39in]{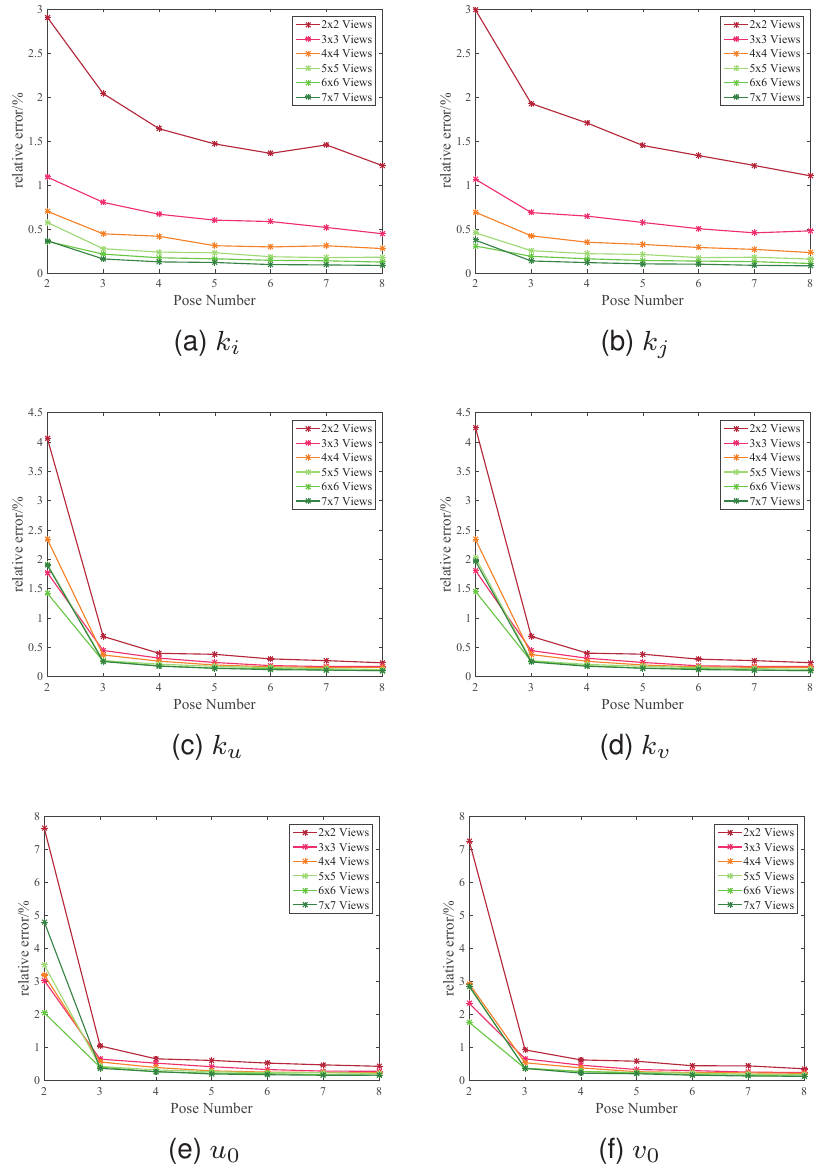}%
\caption{Relative errors of intrinsic parameters on the simulated data with different number of poses and views.}
\label{fig:relative_error_view_pose}
\end{figure}

\begin{figure}[tbp]
\centering
\includegraphics[width=3.39in]{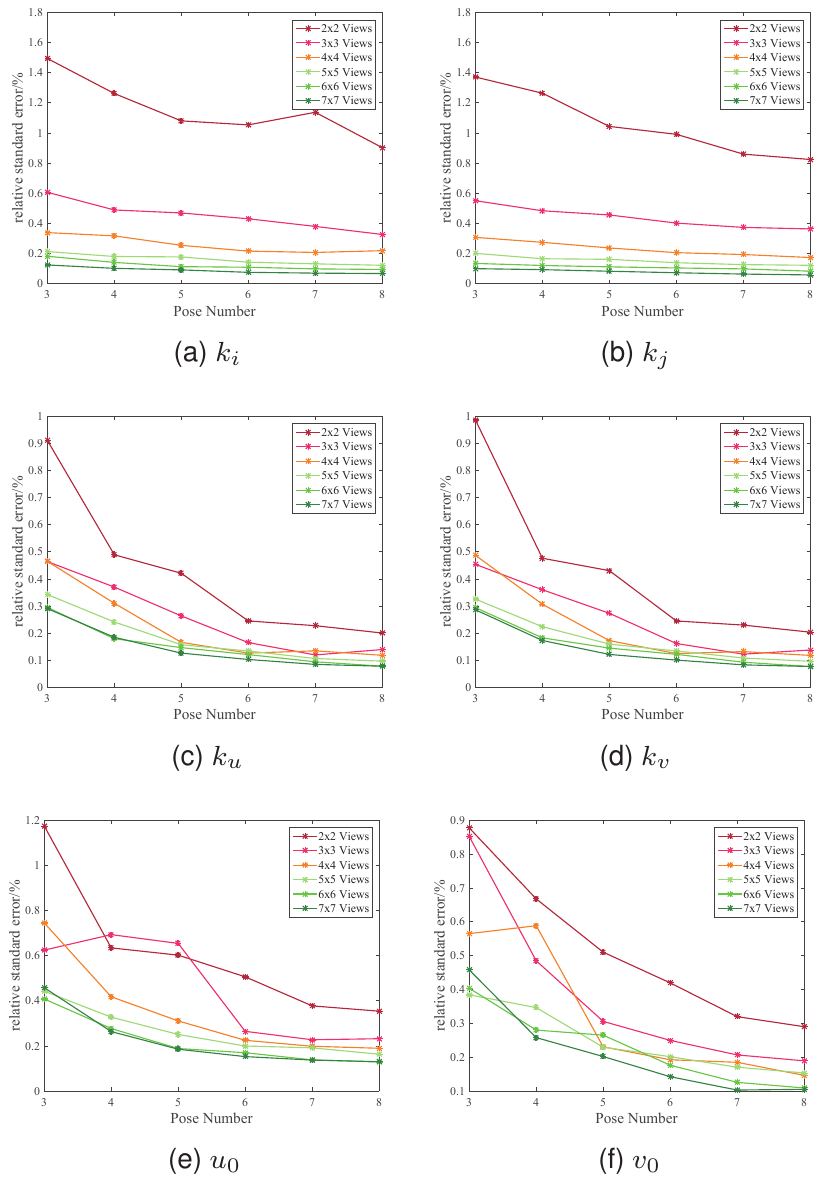}%
\caption{Standard deviations of relative errors of intrinsic parameters on the simulated data with different number of poses and views.}
\label{fig:relative_std_error_view_pose}
\end{figure}

The calibration results with increasing measurements are shown in Fig. \ref{fig:relative_error_view_pose}. We find that the relative errors decrease with the increase in the number of poses. When the number of pose is greater than 2, all the relative errors are within an acceptable level, as summarized in Table\ref{tab:min_max_Errors}. Meanwhile, the errors reduce as the number of views grows once the number of poses is fixed. In particular, when $\# pose\!\!\geq \!3$ and $\# view\!\!\geq\! 4\times4$, all the relative errors are less than $0.5\%$. Furthermore, the standard deviations of relative errors of Fig. \ref{fig:relative_error_view_pose} are shown in Fig. \ref{fig:relative_std_error_view_pose}, from which we can see that standard deviations decrease significantly when the number of pose is greater than 2. Particularly, when $\# pose \!\!\geq\!\!5$ and $\# view\!\!\geq\!\!3\times\!3$, standard deviations keep at a low level stably. The results in Fig. \ref{fig:relative_error_view_pose} and Fig. \ref{fig:relative_std_error_view_pose} have verified the effectiveness of the proposed calibration algorithm.

\begin{table}[t]
\caption{Intrinsic parameter configuration of the simulated light field camera.}
\centering
\scriptsize
\renewcommand\arraystretch{1.}
\begin{tabularx}{.48\textwidth}{CCCCcc}
\toprule
 $k_i$ & $k_j$ & $k_u$ & $k_v$ & $u_0$ & $v_0$\\
\midrule
 2.4000e-04 & 2.5000e-04  & 2.0000e-03 & 1.9000e-03 & -0.3200 & -0.3300\\
\bottomrule
\end{tabularx}
\label{tab:configuration}
\end{table}

\begin{table}[!t]
\caption{Min and Max relative errors of intrinsic parameters (unit: \%) on the simulated data when the number of poses is great than 2.}
\centering
\scriptsize
\renewcommand\arraystretch{1.}
\begin{tabularx}{.48\textwidth}{lCCCCCC}
\toprule
 & $k_i$ & $k_j$ & $k_u$ & $k_v$ & $u_0$ & $v_0$\\
\midrule
Min & 0.0842 & 0.0795  & 0.1019 & 0.1020 & 0.1633 & 0.1295\\
Max & 2.0376 & 1.9238  & 0.6871 & 0.6881 & 1.0511 & 0.9298\\
\bottomrule
\end{tabularx}
\label{tab:min_max_Errors}
\end{table}

\subsubsection{Performance w.r.t the measurement noise}
\begin{figure}[!tbp]
\includegraphics[width=3.39in]{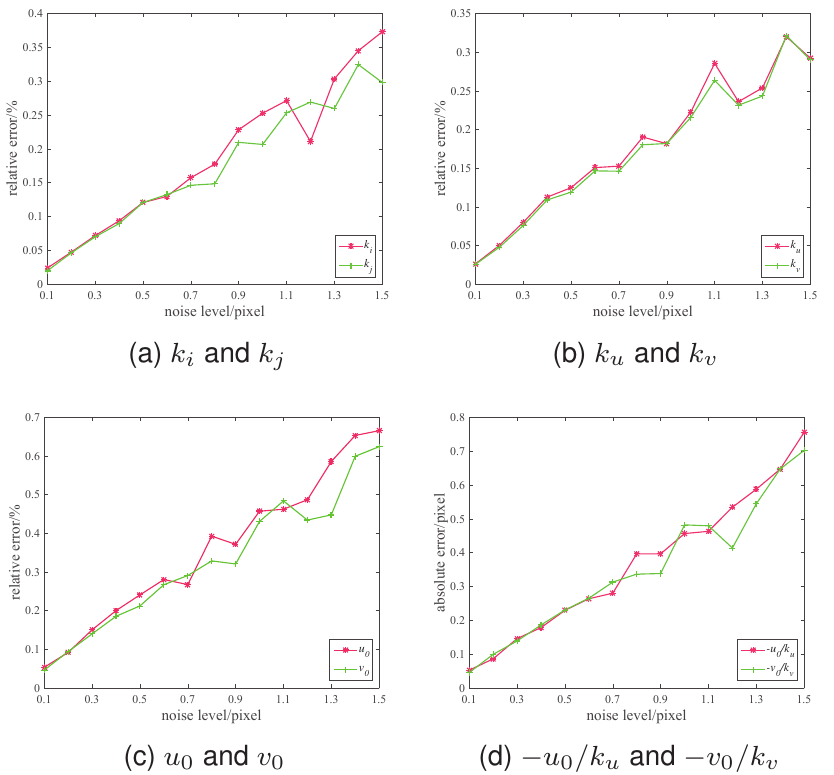}
\caption{Relative errors of intrinsic parameters on the simulated data with different noise levels from 0.1 to 1.5 pixels.}
\label{fig:relative_error_noise}
\end{figure}

Secondly, we employ the measurements of 3 poses and $7\!\times\!7$ views to verify the robustness of calibration algorithm. The rotation angles of 3 poses are $(6^\circ, 28^\circ, -8^\circ)$, $(12^\circ, -10^\circ, 15^\circ)$ and $(-5^\circ, 5^\circ, -27^\circ)$ respectively. Gaussian noise with zero mean and a standard deviation $\sigma$ is added to the projected image points. We vary $\sigma$ from $0.1$ to $1.5$ pixels with a $0.1$ step.
For each noise level, we performed 150 independent trials. The mean results compared with ground truth are shown in Fig. \ref{fig:relative_error_noise}. It demonstrates that the errors increase almost linearly with the noise level. For $\sigma\!=\!0.5$ pixels which is larger than normal noise in practical calibration, the errors of $(k_i, k_j)$ and $(k_u,k_v)$ are less than $0.13\%$. Although the relative error of $(u_0,v_0)$ is $0.24\%$, the absolute error of $(-u_0/k_u,-v_0/k_v)$ is less than $0.23$ pixel (In Eq.(\ref{eq:Decoding}), $u=(x-u_0)/k_u$ and $v=(y-v_0)/k_v$, where $(-u_0/k_u,-v_0/k_v)^\top$ is the principal point of sub-aperture imaging), which further exhibits that the proposed algorithm is robust to higher noise level.

\subsection{Physical camera}\label{subsec:physical}
We also verify the calibration method on real scene light fields captured by conventional and focused light filed cameras. For the conventional light field camera, we use Lytro and Illum to obtain measurements. For the focused one, we use a self-assembly camera according to optical design in Fig. 4a.

\subsubsection{Conventional Light Field Camera}\label{subsubsec:physical1.0}
The sub-aperture images are obtained by the method of Dansereau \textit{et al.} \cite{dansereau2013decoding}. We compare the proposed method in ray re-projection error with state-of-the-arts, including DPW by Dansereau \textit{et al.} \cite{dansereau2013decoding} and BJW by Bok \textit{et al.} \cite{bok2014geometric}.

\begin{figure}[!tbp]
\centering
\includegraphics[width=3.4in]{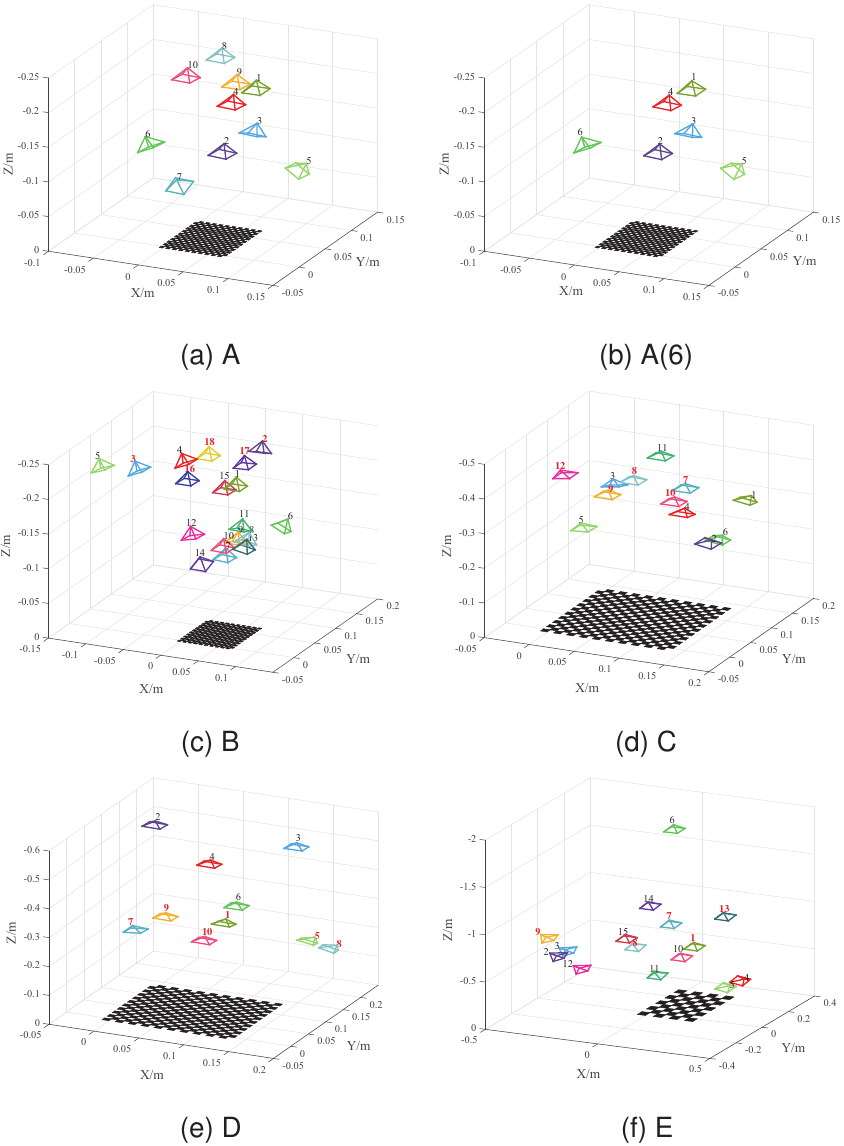}
\caption{Pose estimation results of datasets captured by \cite{dansereau2013decoding}. Light fields used for calibration in Table~\ref{tab:fewer_poses} are indicated with bold red indexes of corresponding camera poses in Figs.~(c-f).  }
\label{fig:pose_ABCDE}
\end{figure}

Firstly, we carry out calibration on the datasets collected with \cite{dansereau2013decoding}. For every different pose, the middle $7\!\times\!7$ sub-apertures are utilized similar to DPW. Table \ref{tab:opt_mm_ABCDE} summarizes the root mean square (RMS) ray re-projection errors of our method and DPW\cite{dansereau2013decoding}. In Table \ref{tab:opt_mm_ABCDE}, the errors of DPW\!\cite{dansereau2013decoding}-1 are taking from the paper directly. The errors of DPW\cite{dansereau2013decoding}-2 are obtained by running their latest released code. On the item of initial, the proposed method provides a smaller ray re-projection error than DPW except on datasets A and B. The result on dataset A performs worse because of bad corner extraction from several poses (\textit{i.e.}, 7$th$, 8$th$, 9$th$ and 10$th$ light field). On the item of optimized, compared with DPW~\cite{dansereau2013decoding} which employs 12 intrinsic parameters, the proposed MPC model only employs a half of parameters but achieves similar performance on ray re-projection errors (the results on datasets A, B and D are better but the results on datasets C and E are worse). Light fields within each dataset are taken over a range of depths and orientations, as shown in Fig. \ref{fig:pose_ABCDE}. The ranges of datasets A, B are $0.25m$ whilst the ranges of datasets C and D do not exceed $0.5m$. Meanwhile, the ranges do not exceed $2m$ in dataset E. Large ranges are reasonable in all datasets only deducing the accuracy in light of distortion model considering the shifted view. This is the main reason why the performance of the proposed method is worse than that of DPW on dataset E. From the dataset A, we select 6 light fields from which the corners are exactly extracted for the proposed method. The ray re-projection error decreases obviously in Table \ref{tab:opt_mm_ABCDE}. Considering the fact that the errors exhibited in DPW are minimized in its own optimization (\textit{i.e.}, ray re-projection error), we additionally evaluate the performance in mean re-projection error of DPW and BJW. As exhibited in Table \ref{tab:opt_pixel_ABCDE}, the errors of the proposed method are obviously smaller than those of DPW and BJW.
{In addition, the calibration with fewer number of poses on the datasets~\cite{dansereau2013decoding} is conducted. For dataset D, we randomly select 6 light fields, and for datasets B, C and E, 5 light fields are randomly selected for calibration. Table~\ref{tab:fewer_poses} summarizes RMS ray re-projection errors and RMS re-projection errors of the proposed method and DPW~\cite{dansereau2013decoding} respectively. In Table~\ref{tab:fewer_poses}, the proposed method achieves smaller errors than DPW. Besides, the calibration results on datasets D and E are obviously improved by reducing the number of poses. We find that smaller range of poses contributes to a performance improvement on datasets D and E, which is shown in Fig.~\ref{fig:pose_ABCDE}.}
Table \ref{tab:intrin_para_ABCDE} lists intrinsic parameter estimation results. The re-projection errors of $7\!\times\!7$ sub-aperture images of B are summarized in Table \ref{tab:sub_aperture_reprojection_B}. The distribution of errors is almost homogeneous. All results have verified the effectiveness of the proposed method.

\begin{table}[!t]
\caption{RMS ray re-projection errors of initial parameter estimation and optimization with distortion rectification (unit: $mm$). The datasets are from \cite{dansereau2013decoding}. The (N) indicates the number of light fields used for calibration among 10 light fields of dataset A. The errors of DPW\!\cite{dansereau2013decoding}-1 are provided by the paper directly, and the errors of DPW\!\cite{dansereau2013decoding}-2 are obtained by running the latest released code.}
\centering
\scriptsize
\renewcommand\arraystretch{1.}
\begin{tabularx}{.48\textwidth}{p{0.8cm}p{1.22cm}CCCCCC}
\toprule
  & & A & A(6) & B & C & D & E\\
\midrule
 & DPW\!\cite{dansereau2013decoding}-1 & 3.2000 & - & 5.0600 & 8.6300 & 5.9200 & 13.8000\\
Initial & DPW\!\cite{dansereau2013decoding}-2 & \textbf{0.5190} & \textbf{0.4229} & \textbf{0.5403} & 0.8832 & 1.1021 & 5.9567\\
 & Ours & 15.3753 & {0.5400} & 0.5952 & \textbf{0.5837} & \textbf{0.7473} & \textbf{2.6235}\\\midrule
 & DPW\!\cite{dansereau2013decoding}-1 & 0.0835 & - & 0.0628 & \textbf{0.1060} & {0.1050} & \textbf{0.3630}\\
Optimized & DPW\!\cite{dansereau2013decoding}-2 & 0.0822 & 0.0903 & 0.0598 & 0.1300 & {0.1149} & 0.3843\\
& Ours & \textbf{0.0810} & \textbf{0.0810} & \textbf{0.0572} & 0.1123 & \textbf{0.1046} & 0.5390\\
\bottomrule
\end{tabularx}
\label{tab:opt_mm_ABCDE}
\end{table}

\begin{table}[!t]
\caption{Mean re-projection errors of optimization with distortion rectification (unit: $pixel$). The results of DPW\!\cite{dansereau2013decoding} are obtained by running their latest released code. The results of BJW\!\cite{Bok2017PAMI} are from their latest paper.} 
\centering
\scriptsize
\renewcommand\arraystretch{1.}
\begin{tabularx}{.48\textwidth}{lCCCCCC}
\toprule
& A & A(6) & B & C & D & E\\
\midrule
DPW\!\cite{dansereau2013decoding} & 0.2284 & 0.3338 & 0.1582 & 0.1948 & 0.1674 & 0.3360\\
BJW\!\cite{Bok2017PAMI} & 0.3736 & - & 0.2589 & - & - & 0.2742\\
Ours & \textbf{0.2200} & \textbf{0.2375} & \textbf{0.1568} & \textbf{0.1752} & \textbf{0.1475} & \textbf{0.2731}\\
\bottomrule
\end{tabularx}
\label{tab:opt_pixel_ABCDE}
\end{table}

\begin{table}[!t]
	\caption{RMS errors of optimization with distortion rectification using fewer poses. The (N) indicates the number of light fields used for calibration.}
	\centering
	\scriptsize
	\begin{tabularx}{.48\textwidth}{lCCCC}
		\toprule
		& \multicolumn{2}{c}{Ray re-projection error} & \multicolumn{2}{c}{Re-projection error}\\
		& \multicolumn{2}{c}{unit: $mm$} & \multicolumn{2}{c}{unit: $pixel$}\\
		\midrule
		& DPW\!\cite{dansereau2013decoding} & Ours & DPW\!\cite{dansereau2013decoding} & Ours\\
		\midrule
		B(5) & 0.0643 & \textbf{0.0622} & 0.2380 & \textbf{0.1458}\\
		C(5) & 0.1260 & \textbf{0.1250} & 0.2323 & \textbf{0.1705}\\
		D(6) & 0.0941 & \textbf{0.0622} & 0.2024 & \textbf{0.1458}\\
		E(5) & 0.2967 & \textbf{0.2888} & 0.3525 & \textbf{0.2049} \\
		\bottomrule
	\end{tabularx}
	\label{tab:fewer_poses}
\end{table}

\begin{table}[!t]
\caption{Intrinsic parameter estimation results of datasets captured by \cite{dansereau2013decoding}.}
\centering
\scriptsize
\renewcommand\arraystretch{1.}
\begin{tabularx}{.48\textwidth}{lCCCCCC}
\toprule
& A & B & C & D & E\\
\midrule
$k_i$ & 2.6998e-04 & 2.7937e-04 & 2.4569e-04 & 2.6833e-04 & 2.3004e-04\\
$k_j$ & 2.7608e-04 & 2.8874e-04 & 2.5359e-04 & 2.6930e-04 & 2.3073e-04\\
$k_u$ & 1.8572e-03 & 1.8357e-03 & 1.8122e-03 & 1.8342e-03 & 1.7585e-03\\
$k_v$ & 1.8692e-03 & 1.8323e-03 & 1.8133e-03 & 1.8352e-03 & 1.7634e-03\\
$u_0$ & -0.3417    & -0.3415    & -0.3550    & -0.3343    & -0.3520\\
$v_0$ & -0.3449    & -0.3344    & -0.3382    & -0.3275    & -0.3615\\
\midrule
$k_1$ & 0.2288     & 0.1829     & 0.1639     & 0.1719     & 0.1612\\
$k_2$ & -0.0928    & 0.0875     & 0.0174     & 0.0213     & -0.0483\\
$k_3$ & -4.5308    & -3.6330    & -3.3591    & -3.5122    & 2.7747\\
$k_4$ & -4.4428    & -3.6064    & -3.3394    & -3.4662    & 2.8320\\
\bottomrule
\end{tabularx}
\label{tab:intrin_para_ABCDE}
\end{table}

\begin{table}[!tbp]
\scriptsize
\caption{RMS re-projection error of sub-apertures in dataset B (uint: $pixel$).}
\centering
\setlength{\tabcolsep}{0.23cm}
\renewcommand\arraystretch{1.}
\begin{tabular}{p{0.27cm}<{\centering}ccccccc}
\toprule
\multirow{2}{*}{\centering $i$} & \multicolumn{7}{c}{$j$}\\
\cmidrule{2-8}
& -3 & -2 & -1 & 0 & 1 & 2 & 3\\
\midrule
-3 & 0.1930 & 0.1820 & 0.1781 & 0.1759 & 0.1759 & 0.1812 & 0.2372\\
-2 & 0.1836 & 0.1763 & 0.1700 & 0.1687 & 0.1718 & 0.1786 & 0.1813\\
-1 & 0.1815 & 0.1724 & 0.1669 & 0.1658 & 0.1692 & 0.1761 & 0.1826\\
 0 & 0.1783 & 0.1731 & 0.1683 & 0.1662 & 0.1713 & 0.1798 & 0.1897\\
 1 & 0.1772 & 0.1733 & 0.1706 & 0.1705 & 0.1748 & 0.1837 & 0.1851\\
 2 & 0.1769 & 0.1761 & 0.1757 & 0.1768 & 0.1809 & 0.1836 & 0.1815\\
 3 & 0.2039 & 0.1746 & 0.1728 & 0.1730 & 0.1798 & 0.1833 & 0.2755\\
\bottomrule
\end{tabular}
\label{tab:sub_aperture_reprojection_B}
\end{table}

\begin{table}[!tbp]
\scriptsize
\caption{RMS ray re-projection errors of initial parameter estimation, optimizations without and with distortion rectification (unit: $mm$).}
\centering
\renewcommand\arraystretch{1.}
\begin{tabularx}{.48\textwidth}{llCCCC}
\toprule
  & & Illum-1 & Illum-2 & Lytro-1 & Lytro-2\\
\midrule
 & DPW\!\cite{dansereau2013decoding} & 0.9355 & 0.6274 & 0.6201 & 0.5057\\
Initial & BJW\!\cite{bok2014geometric} & 1.0765 & 0.8330 & 1.6676 & 1.0201\\
 & Ours & \textbf{0.7104} & \textbf{0.4899} & \textbf{0.3538} & \textbf{0.2364}\\
\midrule
Optimized & DPW\!\cite{dansereau2013decoding} & 0.5909 & 0.4866 & 0.1711 & \textbf{0.1287}\\
without  & BJW\!\cite{bok2014geometric} & - & - & - & - \\
Rectification & Ours & \textbf{0.5654} & \textbf{0.4139} & \textbf{0.1703} & {0.1316}\\
\midrule
Optimized & DPW\!\cite{dansereau2013decoding} & 0.2461 & 0.2497 & 0.1459 & 0.1228\\
with & BJW\!\cite{bok2014geometric} & 0.3966 & 0.3199 & 0.4411 & 0.2673\\
Rectification & Ours & \textbf{0.1404} & \textbf{0.0936} & \textbf{0.1400} & \textbf{0.1124}\\
\bottomrule
\end{tabularx}
\label{tab:opt_216_mm}
\end{table}

\begin{table}[!t]
\scriptsize
\caption{Intrinsic parameter estimation results of our collected datasets.}
\centering
\renewcommand\arraystretch{1}
\begin{tabularx}{.48\textwidth}{p{0.6cm}<{\centering}CCCC}
\toprule
 & Illum-1 & Illum-2 & Lytro-1 & Lytro-2\\
\midrule
$k_i$ & 3.5721e-04 & 2.2464e-04 & 5.9386e-04 & 3.8915e-04 \\
$k_j$ & 3.5455e-04 & 2.3299e-04 & 5.7870e-04 & 3.8247e-04 \\
$k_u$ & 1.4309e-03 & 1.6670e-03 & 9.5083e-04 & 1.3195e-03 \\
$k_v$ & 1.4303e-03 & 1.6657e-03 & 9.4794e-04 & 1.3261e-03 \\
$u_0$ & -0.4565 & -0.5178 & -0.1964 & -0.2775\\
$v_0$ & -0.2827 & -0.3557 & -0.1865 & -0.2521\\
\midrule
$k_1$ & 0.3001 & 0.3562 & -0.4559 & 0.0254\\
$k_2$ & 0.2779 & 0.2595 & 6.8221 & 0.8469 \\
$k_3$ & -1.4109 & -0.6185 & -1.3060 & -2.2441\\
$k_4$ & -1.4204 & -0.8879 & -1.3234 & -2.2684\\
\bottomrule
\end{tabularx}
\label{tab:real_result_intrin_para}
\end{table}

\begin{table}[!t]
\scriptsize
\caption{RMS re-projection error of sub-apertures in dataset Illum-1 (uint: $pixel$).}
\centering
\setlength{\tabcolsep}{0.23cm}
\renewcommand\arraystretch{1.}
\begin{tabular}{p{0.27cm}<{\centering}ccccccc}
\toprule
\multirow{2}{*}{\centering $i$} & \multicolumn{7}{c}{$j$}\\
\cmidrule{2-8}
& -5 & -3 & -1 & 0 & 1 & 3 & 5\\
\midrule
-5 & 0.7880 & 0.2997 & 0.3008 & 0.3041 & 0.3058 & 0.3178 & 0.8070\\
-3 & 0.2992 & 0.3003 & 0.3033 & 0.3025 & 0.3015 & 0.2930 & 0.3077\\
-1 & 0.2988 & 0.3086 & 0.3176 & 0.3182 & 0.3141 & 0.2996 & 0.2827\\
 0 & 0.2942 & 0.3139 & 0.3115 & 0.3058 & 0.3064 & 0.3024 & 0.2772\\
 1 & 0.2934 & 0.3178 & 0.3118 & 0.2963 & 0.3077 & 0.3057 & 0.2784\\
 3 & 0.3002 & 0.2966 & 0.3170 & 0.3093 & 0.3115 & 0.2856 & 0.2843\\
 5 & 0.3283 & 0.2961 & 0.2851 & 0.2854 & 0.2841 & 0.2852 & 0.3102\\
\bottomrule
\end{tabular}
\label{tab:sub_aperture_reprojection_Illum1}
\end{table}

Unlike the core idea of DPW, BJW directly utilizes raw data instead of sub-apertures. However it has a stricter requirement on the acquisition of the calibration board. The data for calibration must be unfocused in order to make the measurements detectable, thus some datasets provided by DPW are incalculable for BJW, just as shown in Table \ref{tab:opt_pixel_ABCDE} (\textit{i.e.} datasets C and D). In order to directly compare with DPW and BJW,
we collect other 4 datasets\footnote{{http://www.npu-cvpg.org/opensource}} using Lytro and Illum cameras. The dataset Illum-1 shoots $9\!\times\!13$ corners with $15.0mm\!\times\!15.0mm$ cells, including 9 poses. The dataset Lytro-1 shoots $8\!\times\!11$ corners with $11.3mm\!\times\!11.3mm$ cells, including 8 poses. The datasets Illum-2 and Lytro-2 shoot $8\!\times\!11$ corners with $8.9mm\!\times\!8.9mm$ cells, including 10 poses. For Illum-1 and Illum-2 datasets, the middle $13\!\times\!13$ views are used ($15\!\times\!15$ views in total). For Lytro-1 and Lytro-2, the middle $7\!\times\!7$ views are used ($9\!\times\!9$ views in total).
Table \ref{tab:opt_216_mm} summarizes the RMS ray re-projection errors compared with DPW and BJW at three calibration stages. As exhibited in Table~\ref{tab:opt_216_mm}, the proposed method obtains smaller ray re-projection errors on the item of initial solution which verified the effectiveness of linear initial solution for both intrinsic and extrinsic parameters. Besides, the proposed method provides similar or even smaller ray re-projection errors on the item of optimization without rectification compared with DPW. It is noticed that the result on dataset Lytro-2 is relatively larger than that of DPW. The main reason is that distortion coefficients $k_3$ and $k_4$ in our model are similar to the elements of the decoding matrix in \cite{dansereau2013decoding}. Considering the fact that MPC model employs less parameters (\textit{i.e.} 6-parameter) than DPW (\textit{i.e.} 12-parameter), the proposed method is competitive with acceptable calibration performance. Further, it is more important that we achieve smaller ray re-projection errors if distortion rectification is introduced in optimization. The ray re-projection errors are encouraging that the proposed method outperforms DPW and BJW. Consequently, the 6-parameter MPC model and 4-parameter distortion model are effective to represent light field cameras.

\begin{figure}[!t]
\centering
\includegraphics[width=3.39in]{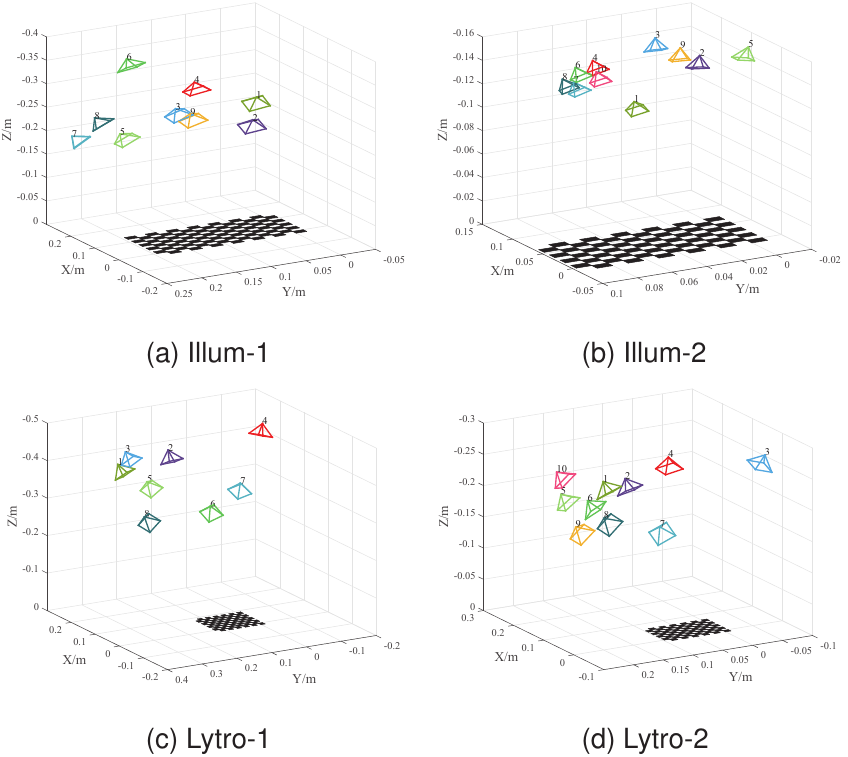}
\caption{Pose estimation results of our collected datasets.}
\label{fig:LFLyrtoIllum_Pose}
\end{figure}
\begin{figure}[!t]
\centering
\includegraphics[width=3.39in]{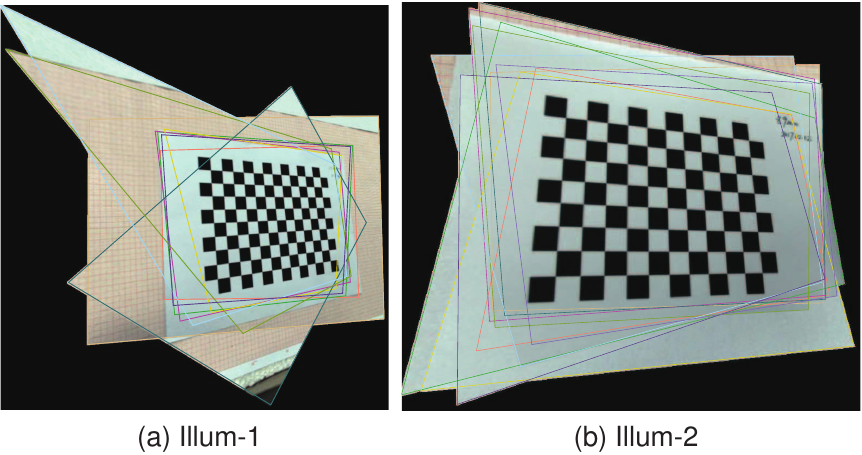}
\caption{The stitching results of Illum-1 and Illum-2 datasets (the first pose is regarded as the reference view).}
\label{fig:Illum_StitchingResults}
\end{figure}
\begin{figure}[!t]
\centering
\includegraphics[width=3.39in]{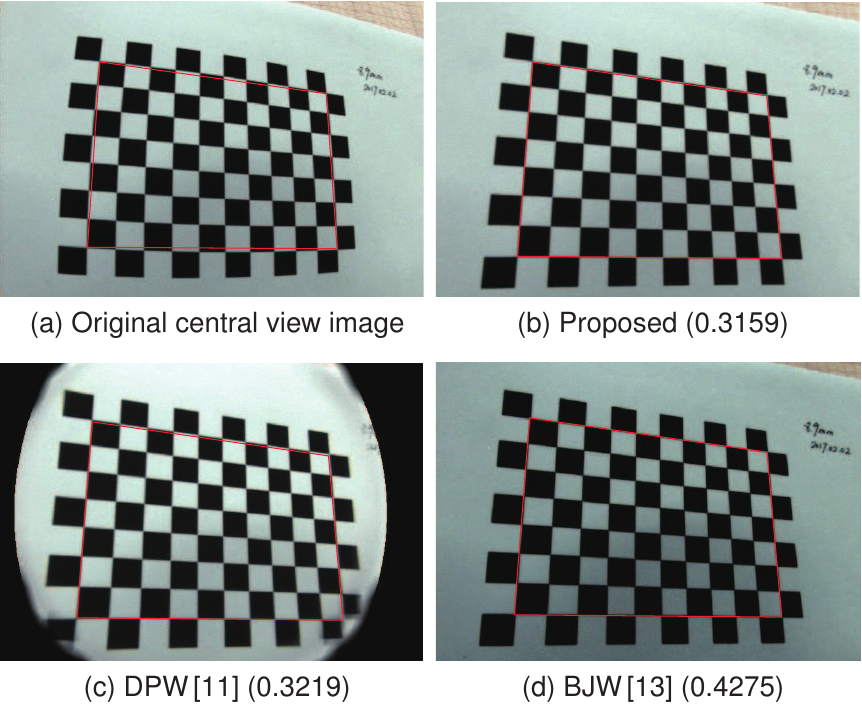}
\caption{The central view sub-aperture and distortion rectification results of first pose light field in Illum-2 dataset. The re-projection error (unit: $pixel$) of central view sub-aperture image is represented in parentheses.}
\label{fig:IllumRectification}
\end{figure}

The reason why we compare ray re-projection errors here is to eliminate differences in camera models. The decoding matrix in \cite{dansereau2013decoding} is similar to Eq.(\ref{eq:Decoding}), except for the non-diagonal elements. The non-zero elements $H_{1,3}$ and $H_{2,4}$ indicate that pixels on the same sub-aperture image have specific relationships among different views. If we calculate the estimated rays by $\mathbf{X}_c$ for the measurement $(s,t,x,y)^\top$, the views may be different. It indicates that there are errors both on the view plane and image plane in \cite{dansereau2013decoding}. As a result, it is not reasonable to compare re-projection error only.

Moreover, the results of intrinsic parameter estimation and pose estimation on our datasets are demonstrated in Table \ref{tab:real_result_intrin_para} and Fig. \ref{fig:LFLyrtoIllum_Pose} respectively. After the calibration process, we measure the RMS re-projection errors of sub-aperture images by utilizing estimated parameters, as shown in Table \ref{tab:sub_aperture_reprojection_Illum1}.
In order to further verify the accuracy of intrinsic and extrinsic parameter estimation, we stitch all other light fields on the first pose, as shown in Fig. \ref{fig:Illum_StitchingResults}, from which we can see all view light fields are registered and stitched very well. Eventually, it is worthy noting that there exists distinct distortion in the Illum camera. In Fig. \ref{fig:IllumRectification}, we show original central view sub-aperture image and rectification results using distortion models of the proposed method, DPW \cite{dansereau2013decoding} and BJW \cite{bok2014geometric} respectively. Since the re-projection error indicates the image distance between a projected point and a rectified one, it can be used to quantify the error of distortion rectification results. In Fig.~\ref{fig:IllumRectification}, we list the RMS re-projection error of central view sub-aperture image using different methods in parentheses, which further verifies that the rectification results of the proposed method are better than those of baseline algorithms.

\begin{figure}[!t]
\centering
\includegraphics[width=3.39in]{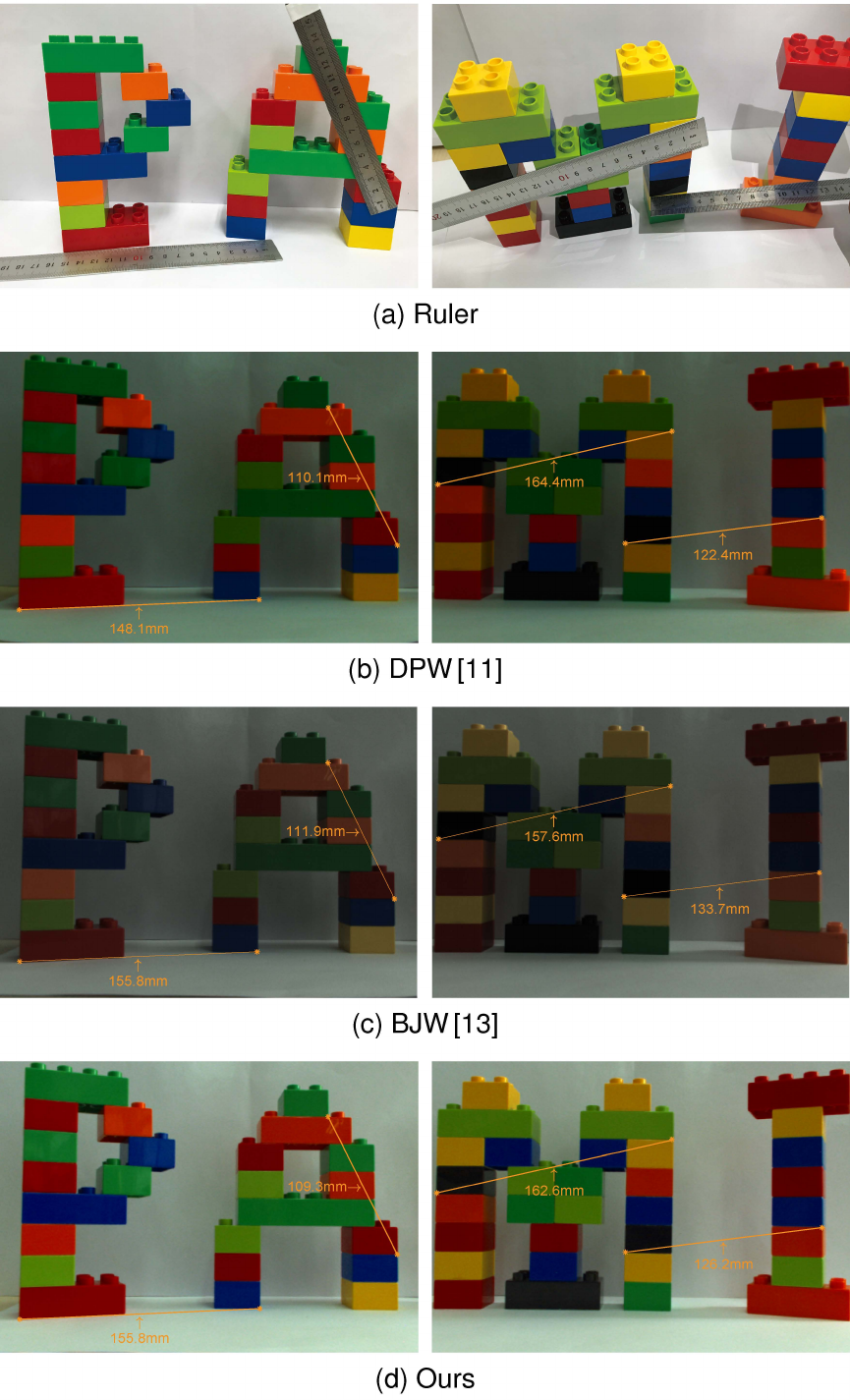}
\caption{The evaluations of light field measurements. (a) shows distances between 3D points measured by rulers. (b), (c) and (d) demonstrate the estimated distances between the reconstructed points after light filed camera calibration with different methods.}
\label{fig:Measurement}
\end{figure}

High-precision calibration is essential in early stages of light field processing pipeline. In order to verify the accuracy of geometric reconstruction of the proposed method compared with baseline methods, we capture two real scene light fields, then reconstruct several {typical} corner points and estimate the distances between them. Fig.~\ref{fig:Measurement} shows reconstruction results on the central view sub-aperture images.  As exhibited in Fig. \ref{fig:Measurement}, the estimated distances between points reconstructed by the proposed method are nearly equal to those measured lengths from real objects by rulers. In addition, Table~\ref{tab:measurement} lists the comparisons of reconstruction results with state-of-the-art methods. The relative errors of reconstruction results demonstrate the performance of our method.

\begin{table}[!t]
	\centering
	\scriptsize
	\caption{Quantitative comparison of different calibration methods (unit: $mm$). The relative error is indicated in parentheses.}
	\setlength{\tabcolsep}{0.06cm}
	\renewcommand\arraystretch{1.}
	\begin{tabularx}{.48\textwidth}{p{1.1cm}<{\centering}CCCC}
		\toprule
		& `P' & `A' & `M' & `I'\\
		\midrule
		Ruler & 155.0 & 108.5 & 163.0 & 128.5 \\
		\midrule
		DPW\!\cite{dansereau2013decoding} & 148.1 (4.45$\%$) & 110.1 (1.47$\%$) & 164.4 (0.86$\%$) & 122.4 (4.75$\%$) \\
		BJW\!\cite{bok2014geometric} & 155.8 (\textbf{0.52}$\%$) & 111.9 (3.13$\%$) & 157.6 (3.31$\%$) & 133.7 (4.05$\%$)\\
		Ours & 155.8 (\textbf{0.52}$\%$) & 109.3 (\textbf{0.74}$\%$) & 162.6 (\textbf{0.25}$\%$) & 126.2 (\textbf{1.79}$\%$)\\
		\bottomrule
	\end{tabularx}
	\label{tab:measurement}
\end{table}

\begin{table}[!t]
	\scriptsize
	\caption{ The running time of initial parameter estimation (unit: $s$).}
	\centering
	\renewcommand\arraystretch{1}
	\begin{tabularx}{.48\textwidth}{lCCCC}
		\toprule
		& Illum-1 & Illum-2 & Lytro-1 & Lytro-2\\
		\midrule
		DPW\!\cite{dansereau2013decoding} & 10.7718  & 12.1735  & 4.3390  & 5.3729 \\
		BJW\!\cite{bok2014geometric} & 20.0266   & 24.2377  & 14.0859 & 15.2629 \\
		Ours & \textbf{0.2359}  & \textbf{0.1878}  & \textbf{0.1263}  & \textbf{0.1428} \\
		\bottomrule
	\end{tabularx}
	\label{tab:time_216}
\end{table}

\begin{figure}[!t]
\centering
\includegraphics[width=3.39in]{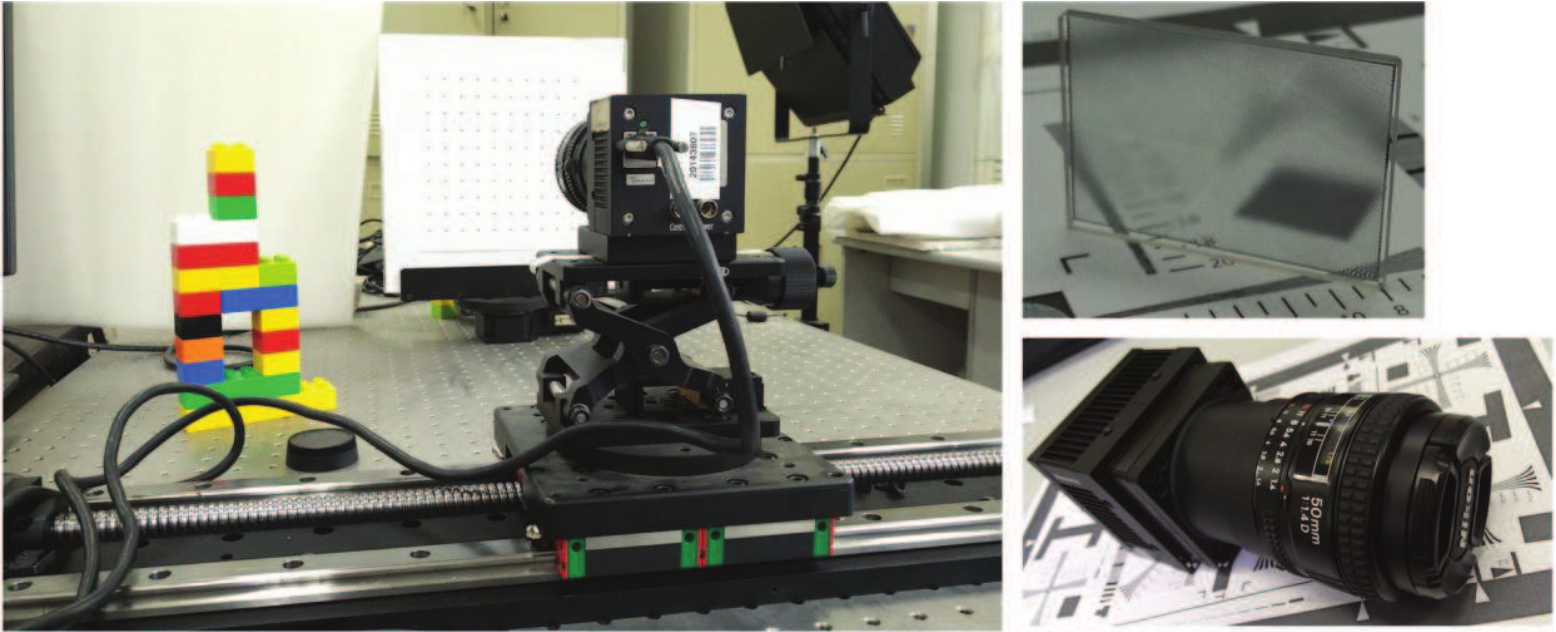}
\caption{The self-assembly focused light field camera and the MLA inside the camera. The light field design pattern is shown in Fig. 4a.}
\label{fig:camera2.0tele}
\end{figure}

\begin{figure*}[!t]
\centering
\includegraphics[width=6.9in]{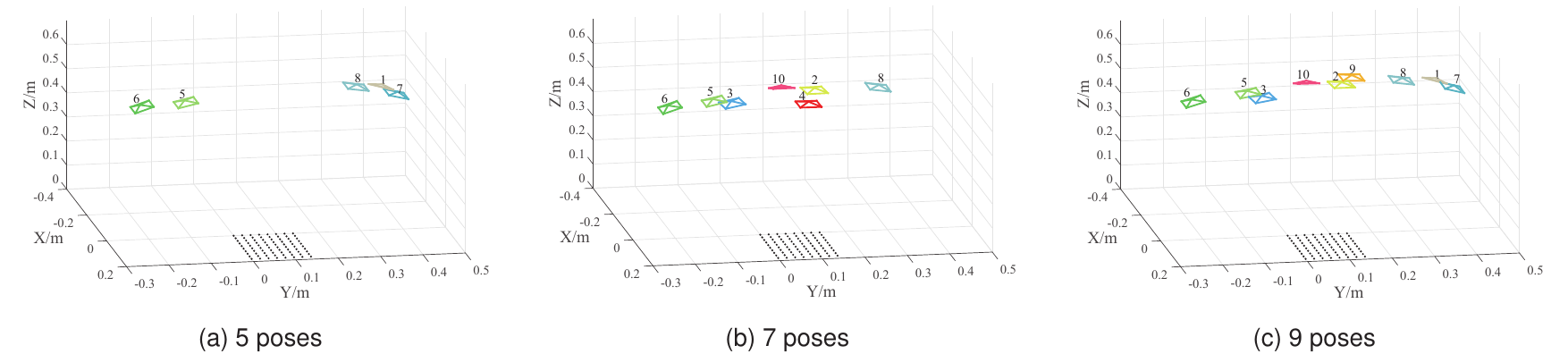}
\caption{Pose estimation results of our self-assembly focused light field camera with different number of poses.}
\label{fig:Plen2_Pose}
\end{figure*}
As mentioned above, since BJW has a stricter requirement on the image of calibration board, some datasets (\textit{i.e.} C and D)~\cite{dansereau2013decoding} are incalculable for BJW. In order to directly compare with DPW and BJW, we utilize our collected datasets to analyze the running time of initial parameter estimation, as illustrated in Table \ref{tab:time_216}. All algorithms are executed in MATLAB 2014 on a desktop computer (2.8GHz CPU and 16G RAM). The proposed method is most efficient comparing to other methods.
Since the computing of SVD in initial linear solution estimation of BJW is time-consuming (\textit{i.e.}  the running time of SVD on four datasets is 18.5277, 22.2501, 13.1004 and 14.2971$s$ respectively), the running time of BJW is shorter than that of DPW if the execution time of SVD is excluded.
The results of running time conform well to our complexity analysis in Sec.\ref{subsec:complexity}.

\subsubsection{Focused Light Field Camera}\label{subsubsec:physical2.0}

We capture data using a self-assembly focused light field camera. The camera and the MLA are shown in Fig. \ref{fig:camera2.0tele}. The camera consists of a GigE camera with a CCD image sensor whose resolution is $4008\!\times\!2672$ with $9 \mu m$ pixel width, a Nikon AF Nikkor f/1.4D F-mount lens with $50 mm$ focal length, and a MLA with $300\mu m$ diameter and $2.726 mm$ focal length in hexagon layout. The light field design pattern is shown in Fig. \ref{fig:camera2.0tele}. The calibration board is $8\!\times\!10$ points with $20mm\!\times\!20mm$ cells. We shoot 10 raw images with different poses of calibration board. Besides, we also collect 9 raw images with real scene and calibration board. All the images are stitched together to generate an enhanced light field with wider FOV.

We carry out calibration using different numbers of light fields from 10 poses (\textit{i.e.}, 5 poses, 7 poses and 9 poses). The estimated intrinsic parameters of our physical camera are listed in Table \ref{tab:real_result_focused_intrin}. The top row shows the estimated results by direct linear initialization. The middle and bottom rows show the optimized results without and with distortion rectification. As shown in Table \ref{tab:real_result_focused_intrin_error}, the RMS re-projection errors are less than 0.75 pixels after the optimization with distortion rectification. Fig. \ref{fig:Plen2_Pose} shows the estimated poses of our physical camera.

\begin{table}[!t]
\caption{Calibration results of intrinsic parameters of our physical camera with different number of poses.}
\centering
\scriptsize
\setlength{\tabcolsep}{0.128cm}
\renewcommand\arraystretch{1.}
\begin{tabular}{p{0.1cm}<{\centering}p{0.7cm}cccccc} 
\toprule
$\#$Pose &  & $k_i$ & $k_j$ & $k_u$ & $k_v$ & $u_0$ & $v_0$\\
\midrule
  &Init.                      & 1.4497e-02 & 1.3755e-02 & 5.3071e-03 & 5.3068e-03 & -0.1239 & -0.2913 \\
5 & $\text{Opt}_{\text{w/o}}$ & 1.4538e-02 & 1.3973e-02 & 4.9733e-03 & 4.9741e-03 & -0.1719 & -0.2965 \\
  & $\text{Opt}_{\text{w}}$   & 1.3710e-02 & 1.2990e-02 & 4.9732e-03 & 4.9740e-03 & -0.1719 & -0.2965 \\
\midrule
  &Init.                      & 1.4160e-02 & 1.1321e-02 & 4.4666e-03 & 4.4723e-03 & -0.1333 & -0.2720 \\
7 & $\text{Opt}_{\text{w/o}}$ & 1.4321e-02 & 1.1900e-02 & 4.6850e-03 & 4.6521e-03 & -0.1732 & -0.2901 \\
  & $\text{Opt}_{\text{w}}$   & 1.3363e-02 & 1.2466e-02 & 4.6850e-03 & 4.6521e-03 & -0.1792 & -0.2901 \\
\midrule
  &Init.                      & 1.4066e-02 & 1.0478e-02 & 5.2007e-03 & 5.1898e-03 & -0.1104 & -0.2675 \\
9 & $\text{Opt}_{\text{w/o}}$ & 1.4046e-02 & 1.0927e-02 & 4.7477e-03 & 4.4723e-03 & -0.1645 & -0.2812 \\
  & $\text{Opt}_{\text{w}}$   & 1.2917e-02 & 1.1741e-02 & 4.7476e-03 & 4.4722e-03 & -0.1645 & -0.2812 \\
\bottomrule
\end{tabular}
\label{tab:real_result_focused_intrin}
\end{table}

\begin{table}[!t]
\caption{The distortion vectors and RMS re-projection errors (unit: $pixel$) of our physical camera with different number of poses.}
\centering
\scriptsize
\renewcommand\arraystretch{1.}
\begin{tabular}{cccccc} 
\toprule
$\#$Pose & $k_1$ & $k_2$ & $k_3$ & $k_4$ & Error\\
\midrule
5 &  5.4550e-04 & 6.0268e-05  & -1.9239e-03 & -2.3802e-03 & 0.6225 \\
7 & -2.4021e-04 & -4.1706e-06 & -2.1438e-03 & 1.5318e-03  & 0.6904 \\
9 & -3.7344e-04 & -7.0466e-06 & -2.5751e-03 & 2.3966e-03  & 0.7363 \\
\bottomrule
\end{tabular}
\label{tab:real_result_focused_intrin_error}
\end{table}

\begin{figure*}[!t]
\centering
\includegraphics[width=6.9in]{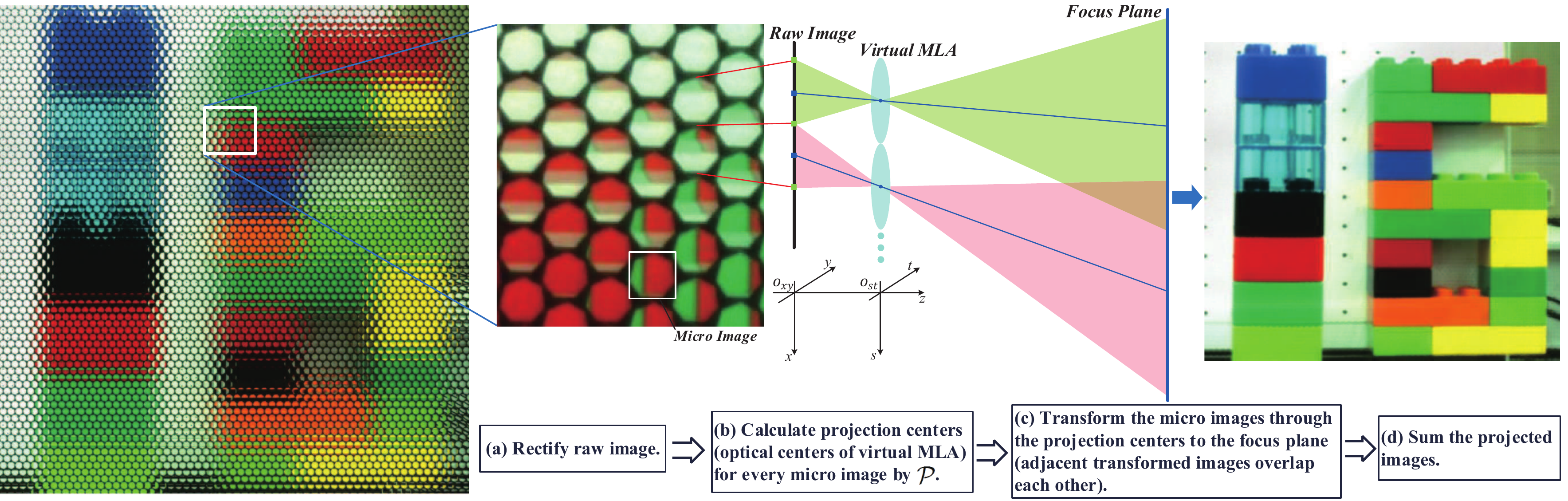}
\caption{The refocus rendering pipeline of a focused light field camera.}
\label{fig:rendering_pipeline}
\end{figure*}

\begin{figure*}[!t]
\centering
\includegraphics[width=6.9in]{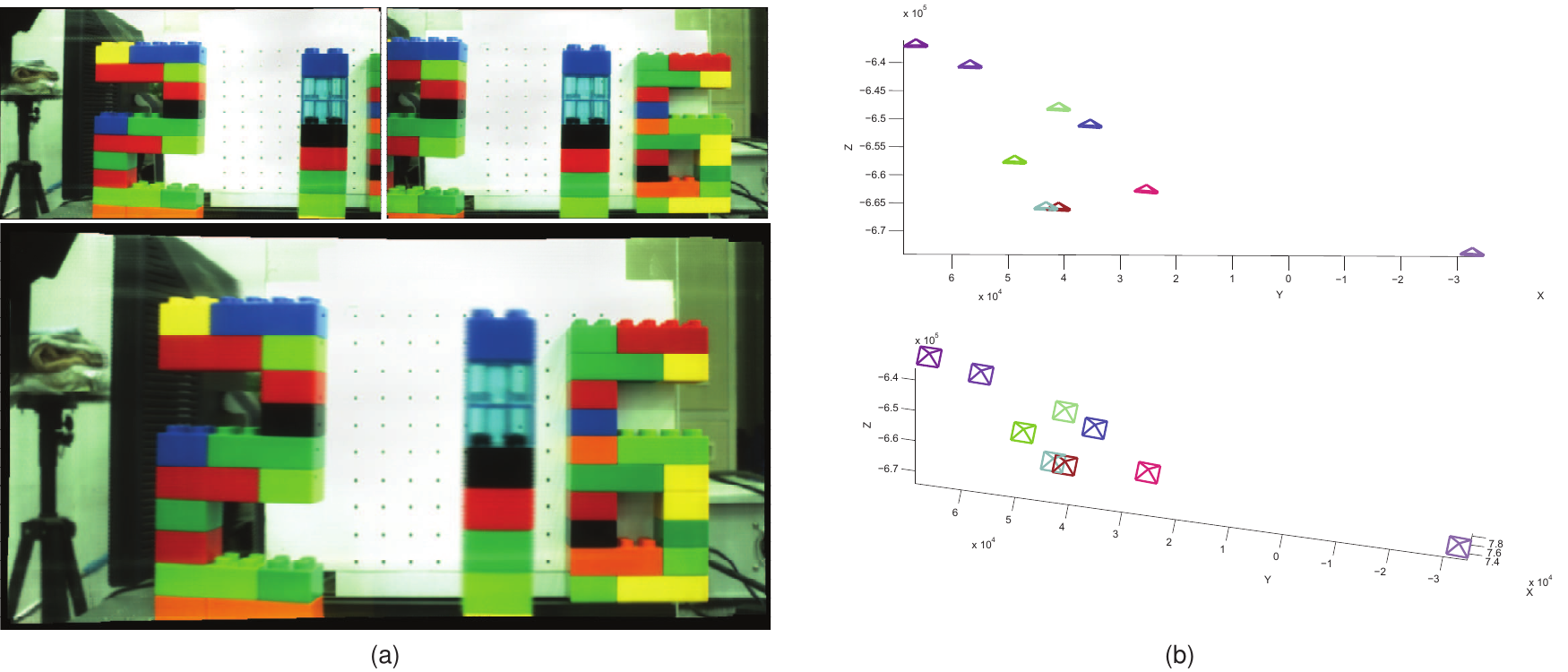}
\caption{Experimental results of light field stitching. (a) shows the rendered images. The top row shows leftmost and rightmost light fields captured by our focused light field camera. The bottom row is refocused result of the stitched light field from 9 light fields. (b) shows the estimated poses of 9 light fields.}
\label{fig:render_image}
\end{figure*}

After the calibration, we render the images by ray tracing using 9 raw images with real scene and calibration board in different poses. The refocus rendering pipeline of a focused light field camera is shown in Fig. \ref{fig:rendering_pipeline}, which includes four key steps. The estimated poses and rendered results are shown in Fig. \ref{fig:render_image}. On the stitched image in Fig. 17a, the dot-array calibration board on the background is focused and the blocks on the foreground are unfocused. The smooth boundary from different poses indicates that our calibration algorithm can estimate exact parameters.

\section{Conclusion}\label{sec:conclusion}
In the paper, we present a multi-projection-center model to parameterize light field and describe light field imaging formation. We deduce the transformations to describe the relationship between 4D rays and 3D scene structure by a projective transformation. Then we verify our light field camera model by intrinsic parameter calibration. We first derive a closed-form solution for initial estimation of intrinsic and extrinsic parameters and then propose a parameter refinement by minimizing the re-projection error. Experiments on conventional light field camera Lytro and Illum and a self-assembly focused light field camera are performed and analyzed extensively. The comparisons with ground truth and state-of-the-art calibration methods have verified the robustness and validity of the proposed model and our calibration method, especially the initialization and optimization with distortion rectification.

In future, we tend to focus on sampling distortion model on the view plane, light field registration and enhancement from un-calibrated cameras for arbitrary scenes, and re-parameterization of 4D light field from arbitrary poses.

\ifCLASSOPTIONcompsoc
  \section*{Acknowledgments}
\else
  \section*{Acknowledgment}
\fi

The authors would like to thank Zhe Ji, Dr. Guoqing Zhou and Dr. Zhaolin Xiao for their helpful suggestions.
{We also thank anonymous reviewers for their valuable feedback.}

\ifCLASSOPTIONcaptionsoff
  \newpage
\fi



%


\bibliographystyle{IEEEtran}
\bibliography{IEEEabrv,egbib}

%

\begin{IEEEbiography}[{\includegraphics[width=1in,height=1.25in,clip,keepaspectratio]{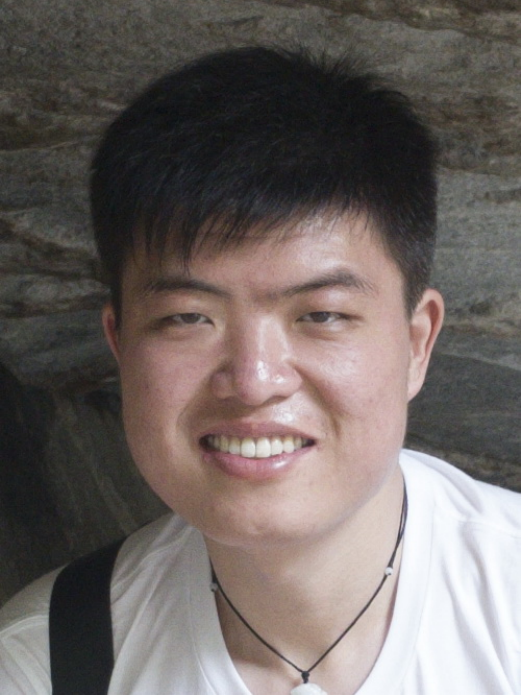}}]{Qi Zhang}  received the B.E. degree in Electronic and Information Engineering from Xi'an University of Architecture and Technology in 2013, and Master degree in Electrical Engineering from Northwestern Polytechnical University in 2015. He is now a Ph.D. candidate at School of Computer Science, Northwestern Polytechnical University. His research interests include computational photography, light field imaging and processing, multiview geometry and applications.
\end{IEEEbiography}

\begin{IEEEbiography}[{\includegraphics[width=1in,height=1.25in,clip,keepaspectratio]{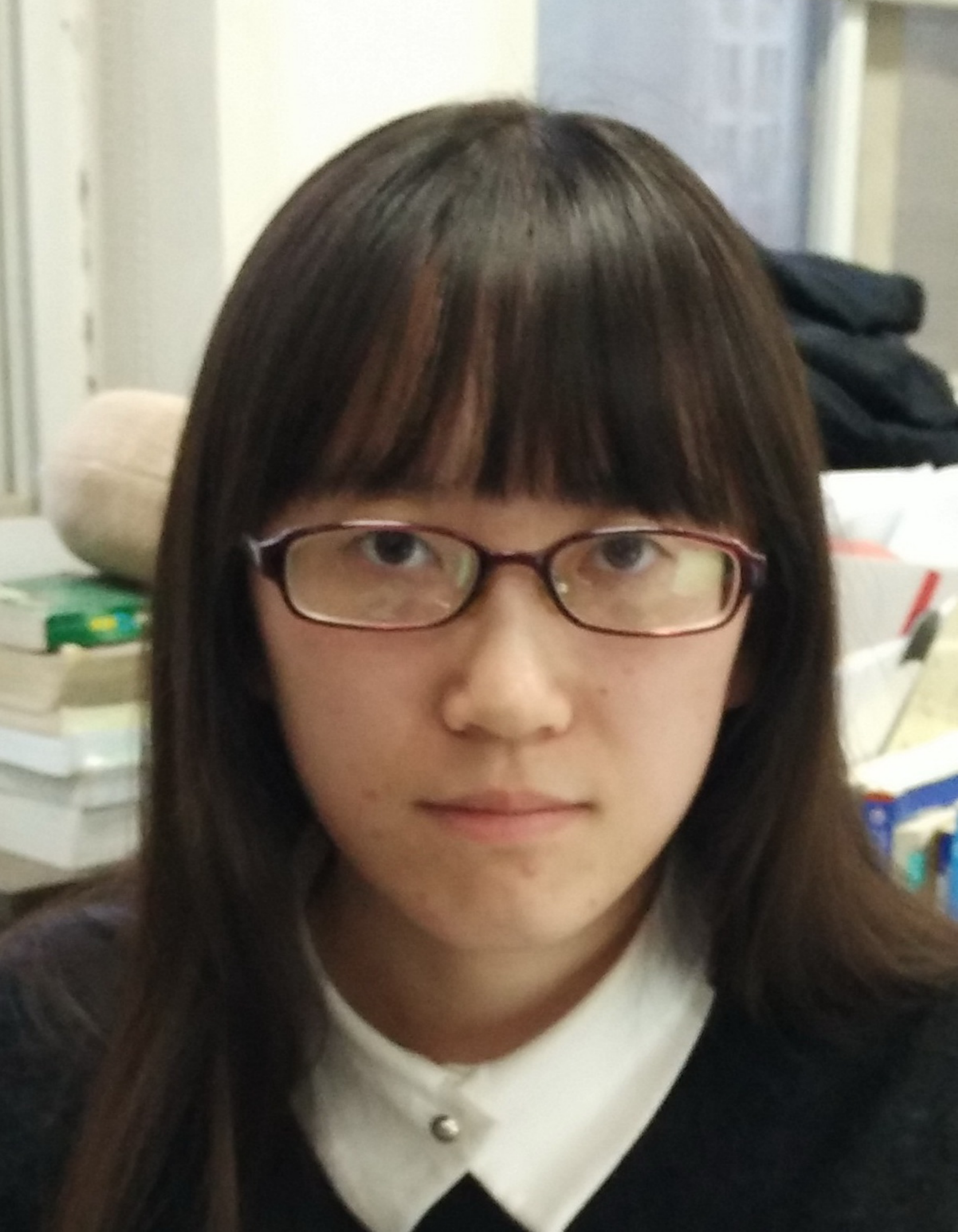}}]{Chunping Zhang} received the B.E. degree and Master degree in Computer Science from School of Computer Science, Northwestern Polytechnical University in 2014 and 2017 respectively. Her research interests include computational photography, light field camera model and calibration, multiview stereo reconstruction.
\end{IEEEbiography}

\begin{IEEEbiography}[{\includegraphics[width=1in,height=1.25in,clip,keepaspectratio]{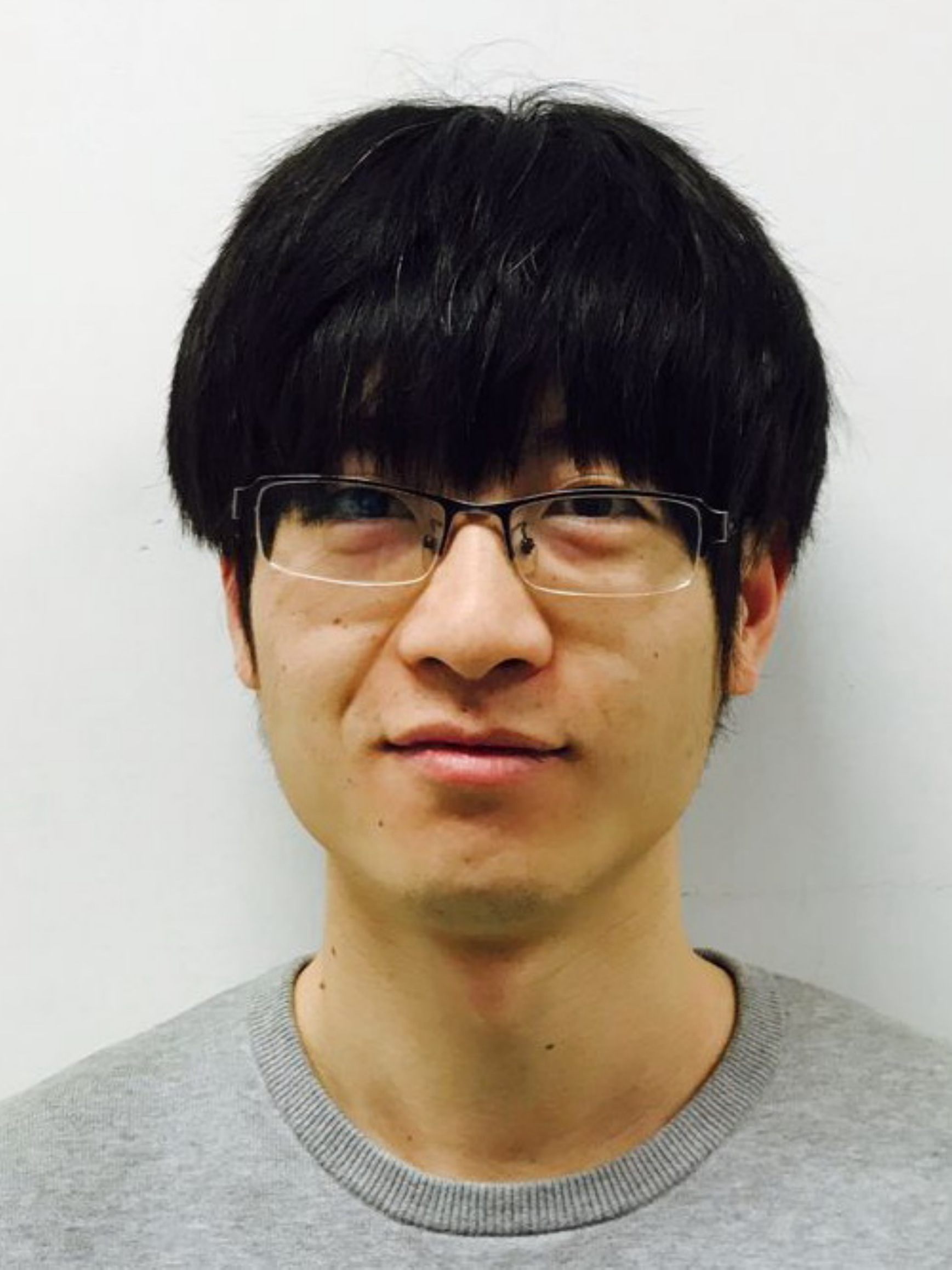}}]{Jinbo Ling} received the B.E. degree in Software Engineering from School of Information Engineering, Chang'an University, in 2016. He is now a postgraduate at School of Computer Science, Northwestern Polytechnical University. His research interests include 3D reconstruction, multiview light field computing and processing.
\end{IEEEbiography}

\begin{IEEEbiography}[{\includegraphics[width=1in,height=1.25in,clip,keepaspectratio]{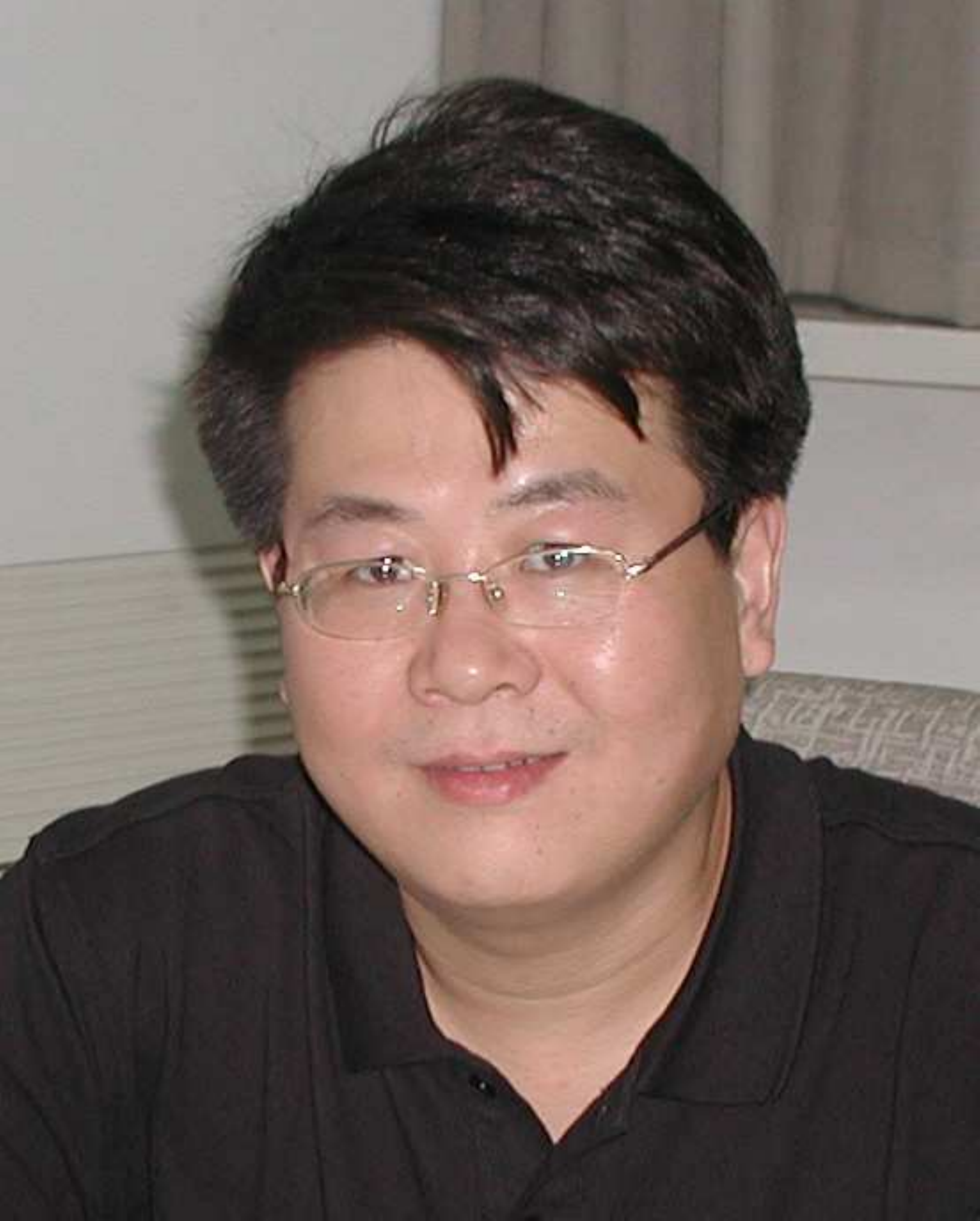}}]{Qing Wang} (M'05) is now a Professor in the School of Computer Science, Northwestern Polytechnical University. He received the B.S. degree from Peking University in 1991. He then joined Northwestern Polytechnical University. In 1997 and 2000 he obtained Master and Ph.D. degrees from Northwestern Polytechnical University. In 2006, he was awarded as outstanding talent program of new century by Ministry of Education, China. He is now a Member of IEEE and ACM. He is also a Senior Member of China Computer Federation (CCF).
He worked as research assistant and research scientist in the Department of Electronic and Information Engineering, the Hong Kong Polytechnic University from 1999 to 2002. He also worked as a visiting scholar in the School of Information Engineering, The University of Sydney, Australia, in 2003 and 2004. In 2009 and 2012, he visited Human Computer Interaction Institute, Carnegie Mellon University, for six months and Department of Computer Science, University of Delaware, for one month.
Prof. Wang's research interests include computer vision and computational photography, such as 3D reconstruction, object detection, tracking and recognition in dynamic environment, light field imaging and processing. He has published more than 100 papers in the international journals and conferences.
\end{IEEEbiography}

\begin{IEEEbiography}[{\includegraphics[width=1in,height=1.25in,clip,keepaspectratio]{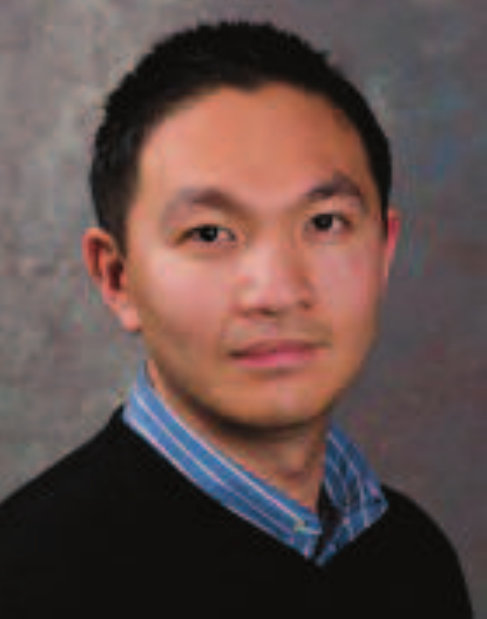}}]{Jingyi Yu} is Director of Virtual Reality and Visual Computing Center in the School of Information Science and Technology at ShanghaiTech University. He received B.S. from Caltech in 2000 and Ph.D. from MIT in 2005. He is also a full professor at the University of Delaware. His research interests span a range of topics in computer vision and computer graphics, especially on computational photography and non-conventional optics and camera designs. He has published over 100 papers at highly refereed conferences and journals including over 50 papers at the premiere conferences and journals CVPR/ICCV/ECCV/TPAMI. He has been granted 10 US patents. His research has been generously supported by the National Science Foundation (NSF), the National Institute of Health (NIH), the Army Research Office (ARO), and the Air Force Office of Scientific Research (AFOSR). He is a recipient of the NSF CAREER Award, the AFOSR YIP Award, and the Outstanding Junior Faculty Award at the University of Delaware. He has previously served as general chair, program chair, and area chair of many international conferences such as ICCV, ICCP and NIPS. He is currently an Associate Editor of IEEE TPAMI, Elsevier CVIU, Springer TVCJ and Springer MVA.
\end{IEEEbiography}



\end{document}